\documentclass[journal]{IEEEtran} 
\usepackage{amsmath,amsfonts}
\usepackage{array}
\usepackage[caption=false,font=normalsize,labelfont=sf,textfont=sf]{subfig}
\usepackage{textcomp}
\usepackage{stfloats}
\usepackage{url}
\usepackage{verbatim}
\usepackage{graphicx}
\usepackage{cite}
\usepackage{hyperref}
\usepackage{xcolor}
\hypersetup{
    colorlinks=true,
    linkcolor=blue, 
    urlcolor=blue,  
    citecolor=blue, 
}

\usepackage{enumitem}
\usepackage{pifont}
\newcommand{\circled}[1]{\ding{\numexpr171+#1\relax}}

\usepackage[ruled]{algorithm2e}
\usepackage{setspace}

\usepackage{adjustbox}

\begin{document}

\title{ColorDynamic: Generalizable, Scalable, Real-time, End-to-end \\Local Planner for Unstructured and Dynamic Environments}


\author{Jinghao Xin, Zhichao Liang, Zihuan Zhang, Peng Wang, and Ning Li, \textit{Member, IEEE}
	\thanks{This work was supported by the National Natural Science Foundation of China (Grant No. 62273230 and 62203302) and the Oceanic Interdisciplinary Program of Shanghai Jiao Tong University (project number SL2023MS011). \textit{(Corresponding authors: Ning Li.)} \par The authors are with the Department of Automation, Shanghai Jiao Tong University, Shanghai 200240, P.R. China, and also with Shanghai Engineering Research Center of Intelligent Control and Management, Shanghai 200240, China, and also with Key Laboratory of System Control and Information Processing, Ministry of Education of China, Shanghai 200240, China (E-mail: xjhzsj2019, zhichaoliang, zhang\_zh317, wangpeng605, ning\_li@sjtu.edu.cn).}
}




\maketitle

\begin{abstract}
Deep Reinforcement Learning (DRL) has demonstrated potential in addressing robotic local planning problems, yet its efficacy remains constrained in highly unstructured and dynamic environments. 
To address these challenges, this study proposes the ColorDynamic framework. 
First, an end-to-end DRL formulation is established, which maps raw sensor data directly to control commands, thereby ensuring compatibility with unstructured environments. 
Under this formulation, a novel network, Transqer, is introduced. The Transqer enables online DRL learning from temporal transitions, substantially enhancing decision-making in dynamic scenarios. 
To facilitate scalable training of Transqer with diverse data, an efficient simulation platform \mbox{E-Sparrow}, along with a data augmentation technique leveraging symmetric invariance, are developed.
Comparative evaluations against state-of-the-art methods, alongside assessments of generalizability, scalability, and real-time performance, were conducted to validate the effectiveness of ColorDynamic. 
Results indicate that our approach achieves a success rate exceeding 90\% while exhibiting real-time capacity (1.2–1.3 ms per planning). 
Additionally, ablation studies were performed to corroborate the contributions of individual components. 
Building on this, the OkayPlan-ColorDynamic (OPCD) navigation system is presented, with simulated and real-world experiments demonstrating its superiority and applicability in complex scenarios. 
The codebase and experimental demonstrations have been open-sourced on our website\footnote{https://github.com/XinJingHao/ColorDynamic} to facilitate reproducibility and further research. 
\end{abstract}


\section{Introduction}
\label{Introduction}
\IEEEPARstart{L}{ocal} planner portrays an instrumental role in robotic navigation systems. Functioning within a hierarchical framework, the local planner controls the robot to follow the path generated by a high-level global planner while dynamically bypassing newly encountered obstacles, enabling robots to execute unmanned tasks such as autonomous driving, environmental exploration, and rescue missions autonomously.
Classical local planning algorithms, exemplified by the Artificial Potential Field (APF)\cite{APF} and the Dynamic Window Approach (DWA)\cite{DWA}, predominantly rely on manually designed rules or parameters, which requires expertise or meticulously engineering, thereby forming an impediment to their widespread applications.

Deep Reinforcement Learning (DRL) has emerged as a promising solution to address these limitations. By enabling robots to proactively learn complex behaviors through interactions with environments, DRL has demonstrated remarkable efficacy in domains such as games\cite{DQN,AlphaGo}, robotics\cite{Color,drlvo}, and large language models\cite{ChatGPT}. Despite these advancements, DRL-based local planners exhibit substantial limitations in highly unstructured and dynamic environments. This constraint arises from three interrelated shortcomings of current researches: 
\mbox{(i) \textit{Pipeline-based Preprocessing}.}
Contemporary DRL-based planners\cite{drlvo,drl-ad,ad-survey} generally rely on multi-stage preprocessing pipelines (e.g., object detection, tracking, and feature fusion) to distill useful environmental features. However, feature extraction from raw sensor data under unstructured environments can be non-trivial. Minor inaccuracies in early pipeline stages propagate cumulatively, often culminating in catastrophic model failures that compromise system robustness. Furthermore, the computational overhead of sequential preprocessing modules impedes real-time performance, rendering such architectures less practical for time-sensitive applications.
\mbox{(ii) \textit{Temporal-Spatial Modeling Dilemma}.}
Existing works\cite{lizhi1,lizhi2,drlvo,pathrl} typically employ Convolutional Neural Networks (CNNs) to process temporally stacked streaming sensor data. While CNNs demonstrate proficiency in spatial feature extraction, their capacity to model temporal dependencies, particularly long-range temporal correlations, remains questionable. Additionally, the localized receptive fields of convolutional operations constrain the model’s ability to integrate global contextual information across spatial and temporal dimensions, leading to CNNs' suboptimal performance under dynamic environments.
\mbox{(iii) \textit{Inefficient Data Accessibility}.}
DRL’s trial-and-error learning paradigm necessitates extensive interaction with environments, posing prohibitive safety and economic risks for real-world training. A prevalent practice within the current research community involves training the planner in Gazebo\cite{lizhi1,lizhi2,pathrl,use-gazebo1,use-gazebo2,use-gazebo3}, a well-established platform for mobile robots simulation. While doable, the effectiveness of Gazebo for DRL training is considerably limited by three key factors: low data throughput, limited data diversity, and an absence of parallelized simulation capabilities. These limitations impede the acquisition of sufficiently diverse training data, thereby restricting the planner’s ability to adapt to real-world conditions and diminishing its generalizability to previously unseen scenarios.

\begin{table*}
\centering
\textbf{Nomenclature}\\
\ \\
\adjustbox{width=\textwidth, center}{
\begin{tabular}{lp{7cm}lp{7cm}}
\hline
\textbf{Symbol} & \textbf{Definition} & \textbf{Symbol} & \textbf{Definition} \\
\hline
$t$ & timestep & $s_t$ & state \\
$o_t$ & partial observation & $a_t$ & action \\
$r_t$ & reward & $\pi$ & policy \\
$\theta$ & parameters of policy & $\gamma$ & discount factor \\
$K(t)$ & kinematic information of the controlled robot & $L(t)$ & lidar scanning result \\
$E(t)$ & encoded temporal lidar data & $T$ & temporal perception length \\
$n$ & number of lidar beams & $ld_n$ & result of the n-th lidar beam \\
$V_l$, $V_a$ & real linear, angular velocity of the robot & $V_l^{max}$, $V_a^{max}$ & maximal linear, angular velocity of the robot \\
$V_l^{ct}$, $V_a^{ct}$ & \textbf{c}urrent \textbf{t}arget \textbf{l}inear, \textbf{a}ngular velocity generated by the agent & $V_l^{rt}$, $V_a^{rt}$ & \textbf{r}eal \textbf{t}arget \textbf{l}inear, \textbf{a}ngular velocity received by the robot \\
$D2T$ & distance from the robot to the target point & $\alpha$ & the robot's relative orientation to the target \\
$r_o, r_a, r_p, r_d$ & oriented-based, action-based, penalty-based, distance-based reward term & $d$ & maximal local planning distance \\
$N_E$ & number of the Transformer encoder layers & $N_P$ & number of MLP layers\\
$N$ & number of vectorized simulation environments & $D$ & feature dimension of the Transformer encoder \\
$Z$ & decision variables of OkayPlan & $F(Z)$ & fitness function of OkayPlan \\
$G(Z)$ & length of global path & $Q(Z)$ & global path's intersections with obstacles \\
$P(Z)$ & global path's intersections with kinematics & $\delta,\sigma,\mu,\nu$ & coefficients of OkayPlan's fitness function \\
$W$ & width of Transqer's linear layers & $H$ & number of heads in Transformer \\
$K,M$ & Kilo, Million & & \\
\hline
\end{tabular}}
\end{table*}

To address these challenges, a novel framework for DRL-based local planning systems is proposed, with the key contributions summarized as follows:
\newpage

\begin{enumerate}
\item{An end-to-end DRL formulation tailored for Local Planning Problems (LPP) is established, which directly maps raw sensor data to low-level control commands. The end-to-end formulation eliminates the need for complex sensor preprocessing, thereby enhancing both the robustness and real-time performance of the planning system in unstructured environments.}
\item{A Transformer-based\cite{transformer} temporal processing model, namely Transqer, is designed to substitute commonly used CNN models and facilitate effective temporal learning in dynamic environments.}
\item{An efficient mobile-robot-oriented simulation platform, E-Sparrow, is developed. Featuring Procedurally Generated Vectorized Diversity (PGVD), the platform enables diverse simulation configurations across multiple dynamic environments through parallel simulation. This strategy contributes to the diversity and throughput of the training data, thereby substantially bolstering the adaptability and generalizability of the trained planner.}
\item{A data augmentation technique based on symmetric invariance for DRL-based robotic systems is engineered to further expand the training data and thus promote the planner's overall performance}
\item{A unified DRL-based local planning framework for unstructured and dynamic environments, termed ColorDynamic, is introduced, integrating the aforementioned techniques.}
\item{Building upon the proposed local planner ColorDynamic and an existing global planner OkayPlan, an advanced navigation system OPCD, featuring high safety and real-time performance, is presented.}
\item{Extensive experiments, in both simulated and real-world environments, are conducted to corroborate the efficacy of the proposed methodologies. Additionally, the full codebase and experimental demonstrations have been open-sourced to promote reproducibility and facilitate further research.}
\end{enumerate}

The remainder of this research is organized as follows. Section II provides a comprehensive review of relevant literature. Section III introduces the proposed ColorDynamic framework and OPCD system. Section IV conducts comparison, evaluation, and ablation studies with respect to the ColorDynamic framework. Section V assesses the OPCD system with existing navigation systems and demonstrates its applicability to real-world scenarios. Finally, Section VI concludes this research.

\section{Related Works}
This section provides a comprehensive review of pertinent studies on the critical components of DRL-based local planners, encompassing temporal perception models, architectures, and simulation platforms. Each subsection concludes with an analysis of their limitations.

\subsection{Temporal Perception in DRL}
\label{TPinDRL}
Temporal perception is critical for DRL agents, enabling the comprehension, prediction, and decision-making required to operate in dynamic environments. Existing methodologies for enhancing temporal perception in DRL can be categorized into three groups: (1) Recurrent Neural Network (RNN)-based, (2) CNN-based, and (3) Transformer-based.
RNN-based methods (e.g., LSTM\cite{use-lstm1,use-lstm2,use-lstm3,use-lstm4} and GRU\cite{use-gru1,use-gru2,use-gru3}) integrate recurrent architectures into policy or value networks to retain historical information. However, RNNs' dependence on initial hidden states restricts their applicability within on-policy DRL algorithms (e.g., A3C\cite{a3c}, PPO\cite{ppo}), which inherently suffer from low sample efficiency.
CNN-based approaches \cite{drlvo,lizhi1,lizhi2,pathrl} address this limitation by reformatting raw sensor data into 2D images, stacking temporal sequences along the depth channel to enable temporal perception. This strategy facilitates the adoption of experience replay\cite{DQN}, ensuring the compatibility with off-policy DRL algorithms thus improving sample efficiency. However, CNNs' efficacy in capturing long-term temporal correlations remains debatable as it is originally designed for spatial feature extraction. 
To address this issue, Transformer-based methods, such as DTQN\cite{DTQN}, conceptualize sequential timesteps as input tokens, leveraging self-attention mechanisms for long-term temporal modeling.
However, Transformer-based models demand substantial training data\cite{scalinglaw,trans_survey1,trans_survey2}, which can be further exacerbated in online DRL settings, where agents must interact extensively with environments to learn useful skills. 
This constraint has forced researchers to resort to offline DRL (e.g., Decision Transformer\cite{decision-transformer}, Trajectory Transformer\cite{traj-transformer}, and Q-Transformer\cite{q-transformer}) as a compromised alternative, which trains agents on pre-collected datasets. Nevertheless, offline methods face challenges in optimality and generalization due to the absence of online exploration. Meanwhile, the preparation of an offline dataset can be labor-intensive. In summary, RNN-based methods confront sample efficiency limitations, CNN-based approaches struggle with long-term temporal modeling, and Transformer-based architectures face challenges in effective learning. Further research is necessary to develop temporally robust DRL frameworks that balance sample efficiency, modeling capacity, and training feasibility.

\subsection{Architectures of DRL-based local planner}
DRL-based local planners are broadly classified into (1) pipeline-based and (2) end-to-end architectures.
Pipeline-based approaches\cite{drlvo,drl-ad,ad-survey} decompose planning into modular components (e.g., object detection, localization, prediction, fusion, planning, and control), offering interpretability and scenario-specific customization. However, the sequential design compromises robustness and real-time performance, as discussed in Section \ref{Introduction}.
In contrast, end-to-end methods\cite{lizhi1,lizhi2,Color,visual-end2end} map the raw sensor data directly to low-level control commands, making them more integrated, thus overcoming the flaws of pipeline-based approaches. Whereas, the design and training of end-to-end architectures are non-trivial, as extracting high-level features directly from noisy sensor data is challenging. 
Consequently, existing approaches typically rely on easier-to-train models (e.g., RNN, CNN), leaving a research gap between end-to-end architectures and advanced temporal models (e.g., Transformer). Hence, further works are necessitated to explore their effective integration and form a more powerful architecture for DRL-based local planners.

\subsection{Simulation Platforms for DRL}
Simulation platforms are indispensable for DRL, as they not only circumvent the prohibitive costs of real-world training but also accelerate the training process. Widely adopted platforms include Robotics Gym\cite{robogym}, PyBullet\cite{PyBullet}, Webots\cite{Webots}, Gazebo\cite{gazebo}, and Sparrow\cite{Color}. Developed by OpenAI, Robotics Gym is a simulation platform based on the MuJoCo physics engine, providing a suite of continuous control tasks and serving as critical tools for robotic control research. 
PyBullet, a high-fidelity physics simulation library based on the Bullet physics engine, is commonly employed in DRL scenarios requiring precise physical interactions, such as robotic manipulation, locomotion, and collision detection. However, despite their advanced capabilities, these platforms are primarily designed for articulated robots and thus cannot adequately support the training of local planners for wheeled robots.
Webots and Gazebo, developed by Cyberbotics and Open Robotics respectively, are two simulation platforms well-suited for wheeled robots. Both support diverse robotic platforms and sensors while providing extensive tools for modeling, simulating, and analyzing robotic behaviors across various environments. This enables researchers to explore different configurations and scenarios in simulation. Nevertheless, due to their high computational demands and lack of parallel simulation capabilities, these platforms exhibit limitations in data throughput and diversity. 
To address these challenges, \cite{Color} proposed the Sparrow simulator. Featuring vectorized diversity, Sparrow enables simultaneous simulations of diversified setups (e.g., environmental layouts, sensor noise, and robot kinematics) across multiple vectorized environments. This functionality enhances both data diversity and throughput, thereby ameliorating the DRL training significantly. 
However, Sparrow currently supports only static environment simulations, restricting its applicability in more complex and realistic scenarios involving dynamic obstacles. Furthermore, its diverse training maps require manual preparation, substantially increasing labor costs and diminishing the automaticity of DRL training. Therefore, further research is necessary to develop a more comprehensive and advanced platform apposite for DRL-based local planning.

\section{Methodology}
This section introduces the ColorDynamic. We commence with the end-to-end problem formulation, followed by the E-Sparrow simulator, the Transqer networks, and the SI data augmentation technique.

\subsection{End-to-end Formulation for LPP}
In this section, we introduce our end-to-end formulation for the LPP, which maps raw temporal sensor data directly to the low-level control command within the DRL diagram.

\subsubsection{Problem Formulation}
Due to perceptual limitations (e.g., constrained lidar scanning range), the agent cannot fully observe the environment. Consequently, the LPP is formulated as a Partially Observable Markov Decision Process (POMDP) to facilitate the deployment of DRL. 
Let $s_t$ denote the true state of the environment at timestep $t$. The agent then receives a partial observation $o_t$ of $s_t$ through its sensors and takes an action $a_t$ based on its policy $\pi(o_t;\theta)$, where $\theta$ represents the policy's parameters. 
Subsequently, the environment provides a reward $r_t$ and transitions to the next state $s_{t+1}$. The objective of the agent is to maximize the expected sum of discounted rewards $\mathbb{E}_\pi\left[\sum_{t=0}^{+\infty} \gamma^t r_t\right]$, where $\gamma \in[0,1]$ is the discount factor.

\subsubsection{Observation}
The primary objective of the agent within the LPP is to control the robot to reach its target point as quickly as possible while avoiding collisions with obstacles along the route, necessitating effective perception of the surrounding environment. Although sensors such as RGB and depth cameras provide abundant feature-rich data, their employment in the LPP, however, encounters substantial challenges:
(i) Limited generalization capability: feature-rich sensors are prone to overfitting to environmental specifics\cite{cnn_overfit} (e.g., colors, textures, and shapes), leading to degraded performance in unseen or dynamic environments.
(ii) Low computational efficiency: the use of feature-rich sensors requires intricate preprocessing pipelines\cite{drlvo}, which not only demand additional human effort but also impose significant computational costs, posing a critical challenge for real-time decision-making.

Our prior research\cite{Color} demonstrated that single-stream lidar is highly effective for the LPP of mobile robots, as it provides essential collision-avoidance information, such as obstacle distances, while excluding extraneous data that may lead to overfitting. Additionally, the inherent simplicity of single-stream lidar data facilitates direct integration into neural networks, enabling end-to-end feature extraction and decision-making. This eliminates the need for hand-crafted preprocessing, consequently enhancing real-time performance. 
Moreover, this choice reduces redundant sensors, contributing to the simplicity, affordability, and energy efficiency of the robot system. 
Consequently, this study employs single-stream lidar as the perception sensor, with the agent's observation defined accordingly:
\begin{equation}
	\label{Ot}
	o_t = [K(t),L(t),L(t-1),...,L(t-T+1)]
\end{equation}
\noindent where $T$ is the temporal perception length, $L(t)$ is the lidar scanning result at timestep $t$:

\begin{equation}
	\label{Lt}
	L(t) = [ld_1, ld_2, ld_3, \cdots, ld_n]
\end{equation}
\noindent where $n$ is the number of lidar beams.
$K(t)$  in Eq. (\ref{Ot}) denotes the kinematic information of the mobile robot at timestep $t$:

\begin{equation}
	\label{Kt}
	K(t)=[V_l^{ct}, V_a^{ct}, V_l^{rt}, V_a^{rt}, D2T, \alpha, V_l, V_a]
\end{equation}

\noindent where $V_l^{ct}$ and $V_a^{ct}$ are the \textbf{c}urrent \textbf{t}arget \textbf{l}inear and \textbf{a}ngular velocity generated by the agent, respectively; $V_l^{rt}$ and $V_a^{rt}$ are the \textbf{r}eal \textbf{t}arget \textbf{l}inear and \textbf{a}ngular velocity received by the robot, respectively; $D2T$ is the distance from the robot to the target; $\alpha$ is the robot's relative orientation to the target; $V_l$ and $V_a$ are the real \textbf{l}inear and \textbf{a}ngular velocity of the robot, respectively.

\begin{figure}[t]
	\centering
	\includegraphics[width=0.31\textwidth]{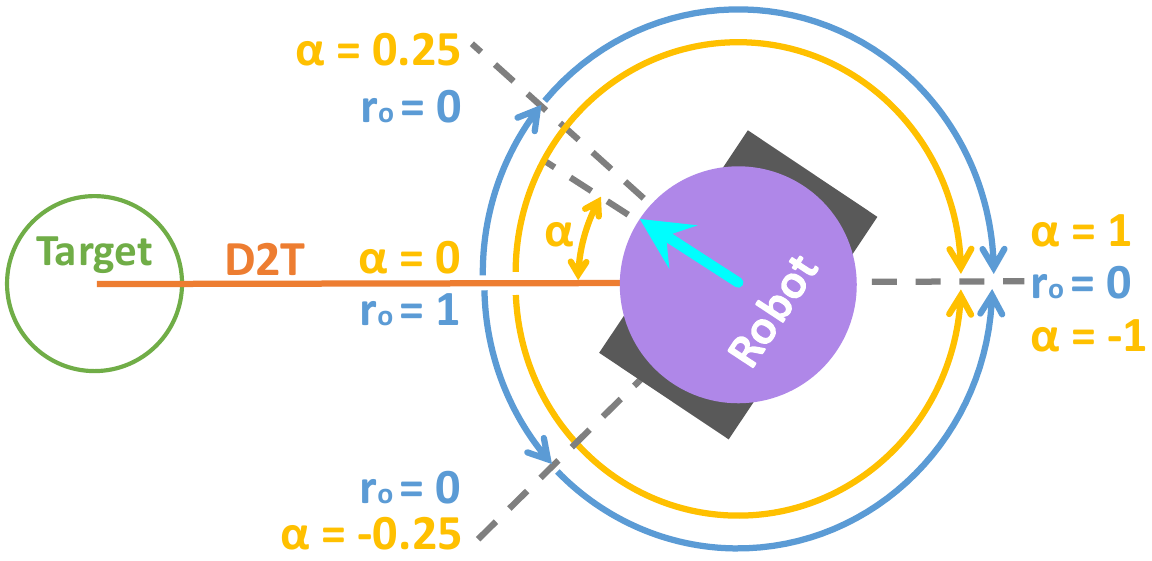}
	\caption{Illustration of $D2T$, $\alpha$, and $r_o$.}
	\label{OR_definition}
\end{figure}

We distinguish between $[V_l^{ct}, V_a^{ct}]$ and $[V_l^{rt}, V_a^{rt}]$ because, in robotic systems, the target velocity generated by the agent and the target velocity received by the robot at a specific timestep are often disparate due to transmission delays. Similarly, we differentiate between $[V_l^{rt}, V_a^{rt}]$ and $[V_l, V_a]$ as the robot's actual operational velocity cannot change abruptly due to physical constraints. Consequently, the target velocity received by the robot and its actual operational velocity at a given timestep are also distinct. We contend that such configurations can equip the agent with more accurate information about the robot, thereby enhancing its performance in complex scenarios. Fig. \ref{OR_definition} provides an illustration of $D2T$ and $\alpha$. These parameters describe the relative position between the robot and the target, which is indispensable for the LPP.

\subsubsection{Action}
The agent utilizes seven discrete actions to generate the target velocity $[V_l^{ct}, V_a^{ct}]$ for the robot:

\begin{enumerate}[label=\circled{\arabic*}]
	\item{\textit{Turn Left}: [0.1 m/s, 2 rad/s]}
	\item{\textit{Turn Left and Move Forward}:[0.5 m/s, 2 rad/s]}
	\item{\textit{Move Forward}: [0.5 m/s, 0 rad/s]}
	\item{\textit{Turn Right and Move Forward}: [0.5 m/s, -2 rad/s]}
	\item{\textit{Turn Right}: [0.1 m/s, -2 rad/s]}
	\item{\textit{Move Backward}: [-0.5 m/s, 0 rad/s]}
	\item{\textit{Slow Down}: [0.05 m/s, 0 rad/s]}
\end{enumerate}

We choose the discrete action space as we find it shows better stability and convergence properties in the LPP than the continuous action space. It is crucial to emphasize that the discrete action setup does not compromise the robot’s movement continuity, as it only controls its target velocity. The actual control signal for the robot’s actuator is generated by a low-level velocity tracker, namely the Proportional-Integral-Derivative\cite{pid1,pid2} controller, in a continuous manner. 

\subsubsection{Reward}
The reward function for the LPP is given by:

\begin{enumerate}[label=\circled{\arabic*}]
	\item $r=200$, if the robot reaches the target point.
	\item $r=-200$, if the robot collides with obstacles or exceeds the maximal local planning distance $d$.
	\item $r=r_o \cdot r_a + 0.5r_p + 0.5r_d -0.5$, otherwise.
\end{enumerate}

The episode will be terminated when reward cases \circled{1} and \circled{2} occur.
Here, $r_o$ and $r_a$ are orientation-based and action-based rewards, respectively. These two reward terms encourage the robot to move rapidly toward the target point. An illustration of $r_o$ is provided in Fig. \ref{OR_definition}, where
\begin{equation}
	\label{ro}
	r_o = 1-\min(0.25, |\alpha|)/0.25
\end{equation}

$r_a=1$ if the action generated by the agent is \textit{Move Forward}, otherwise $r_a=0$.
$r_p$ is a penalty-based reward that discourages backward movement of the robot. $r_p=-1$ if the action is \textit{Move Backward}, otherwise $r_p=0$. 
$r_d$ denotes the distance-based reward, designed to incentivize robot locomotion towards the target point, and is defined as:
\begin{equation}
	\label{rd}
	r_d=\frac{D2T(t-1)-D2T(t)}{V^{max}_l}
\end{equation}

\noindent where $V^{max}_l$ is the maximal linear velocity of the robot.
Note that the reward function is manually engineered, and its coefficients have been tuned in our pre-experiment.

\subsection{Simulation Platform}
Simulation environments have been instrumental in accelerating DRL research for robotics, allowing for safe and efficient training of complex behaviors without the need for costly and time-consuming real-world experiments. \mbox{Gazebo\cite{gazebo}}, the official simulation platform of the Robot Operating System (ROS)\cite{ros}, has been widely adopted for training DRL-based mobile robot systems \cite{CBSS,TRO-Gazebo,MRPD_Gazebo,gazebo_36hours,drlvo} due to its high-fidelity simulation of the physical environment. However, the computational demands and the lack of parallel simulation support result in Gazebo's inefficiencies in both computational resource utilization and training time. For instance, \cite{drlvo} utilizes eight Tesla V100 GPUs to train a DRL-based local planner using Gazebo, while \cite{gazebo_36hours} reports 36 hours of Gazebo simulation to train a DRL-based navigator. This inefficiency substantially impedes the development of DRL-based mobile robot systems.

\begin{figure*}[t]
	\centering
	\subfloat[]{\includegraphics[width=0.15\textwidth]{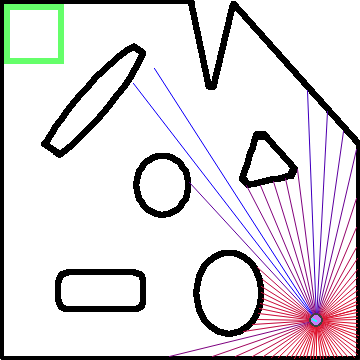}
		\label{S0}}
	\hfil
	\subfloat[]{\includegraphics[width=0.15\textwidth]{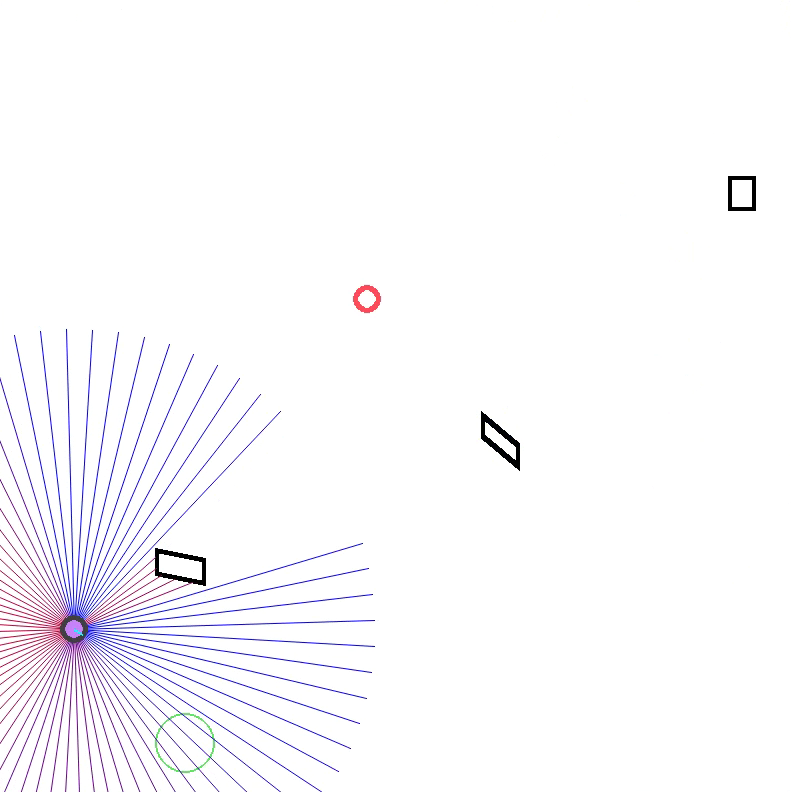}
		\label{S1}}
	\hfil
	\subfloat[]{\includegraphics[width=0.15\textwidth]{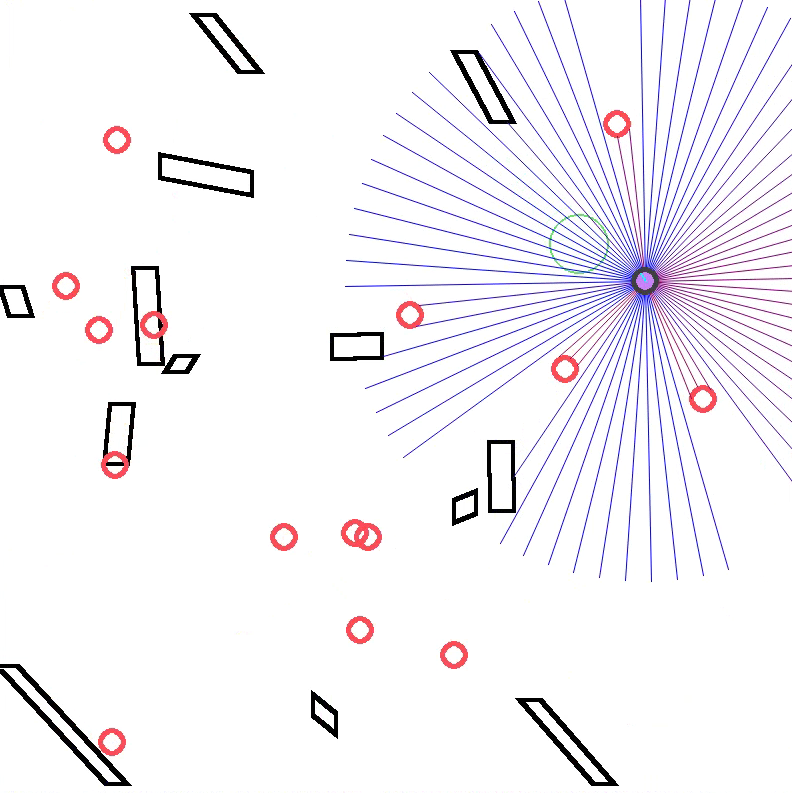}
		\label{S2}}
	\hfil
	\subfloat[]{\includegraphics[width=0.15\textwidth]{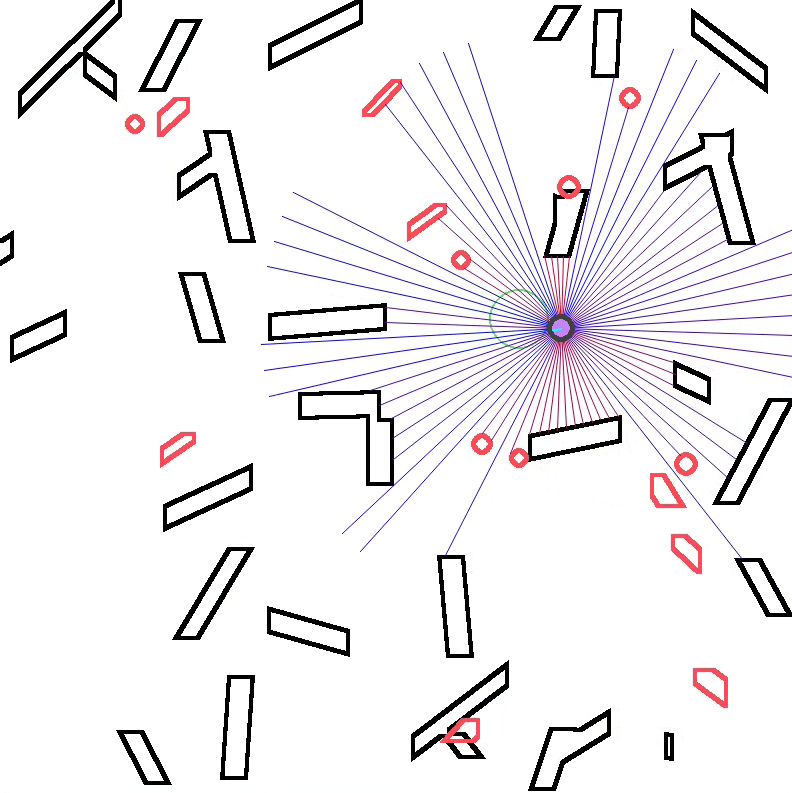}
		\label{S3}}
	\hfil
	\subfloat[]{\includegraphics[width=0.15\textwidth]{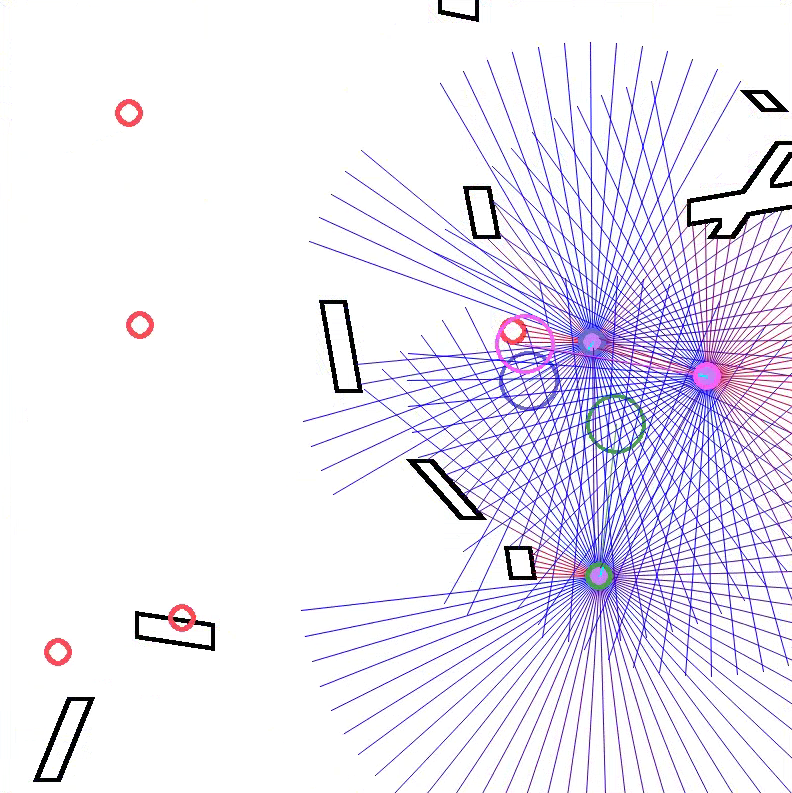}
		\label{S4}}
	\hfil
	\subfloat[]{\includegraphics[width=0.15\textwidth]{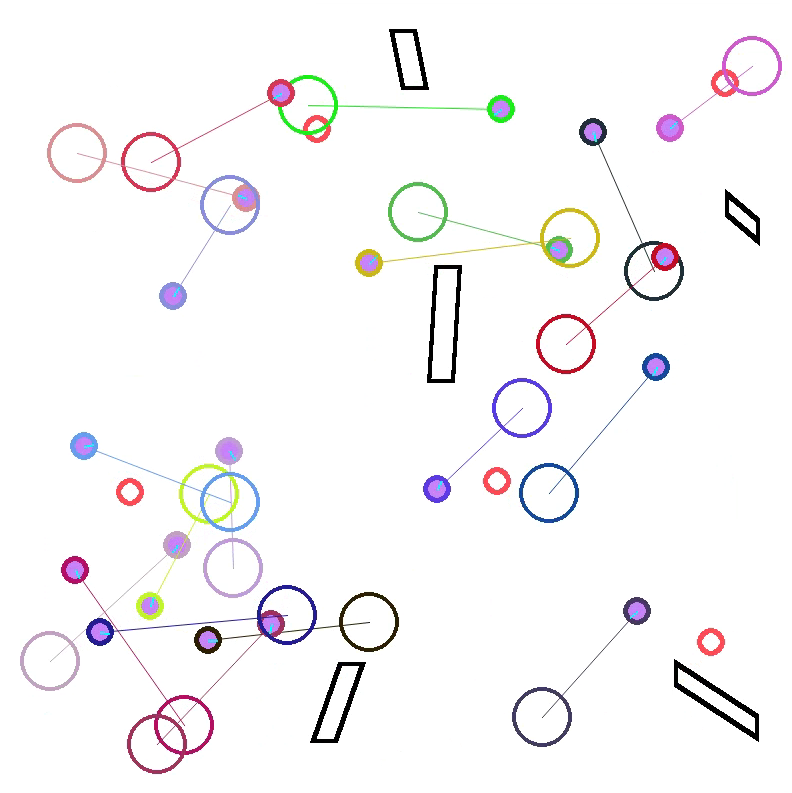}
		\label{S5}}
	\hfil
	\caption{A comparison between Sparrow (a) and E-Sparrow (b)$\sim$(f). The purple circles represent the robots, while the blue fan-shaped lines indicate the lidar beams. The black lumps and red circles denote static and dynamic obstacles, respectively. The green-framed regions highlight the target points. In the multi-robot scenarios (e) and (f), to distinguish the target points of different robots, their colors are aligned with robots' external contours. It is noteworthy that in E-Sparrow, obstacles are randomly generated by the program, and dynamic obstacles move randomly within the environment.}
	\label{SparrowCompare}
\end{figure*}

To address this challenge, our previous work\cite{Color} introduced a lightweight and efficient simulator, Sparrow\footnote{https://github.com/XinJingHao/Sparrow-V1}, specifically designed for mobile robot systems. Sparrow is characterized by its vectorized diversity, enabling simultaneous simulations of diversified setups (e.g., environmental layouts, sensor noise, and robot kinematics) across multiple vectorized environments. This capability enhances both the data diversity and throughput. With a desktop-class computer, Sparrow can yield a local planner within one hour of simulation training\cite{Color}. Furthermore, the simulation-trained local planner can be applied directly to real-world scenarios without further adaptation, owing to the strong generalization capabilities facilitated by vectorized diversity.

Despite these advancements, the Sparrow simulator presented three notable limitations. First, it supported only static obstacle simulations. Second, it was restricted to single-agent simulation. Third, the creation of diverse training maps required manual effort, resulting in substantial human intervention. To address these limitations, this research introduces an enhanced version of the Sparrow simulator, termed E-Sparrow. As illustrated in Fig. \ref{SparrowCompare}, E-Sparrow now incorporates dynamic obstacle and multi-agent simulation capabilities. More importantly, it features Procedurally Generated Vectorized Diversity (PGVD) for training maps, automating the creation of diverse training environments. These enhancements provide E-Sparrow with a more comprehensive and powerful simulation ability, enabling the autonomous generation of unlimited diverse training samples with high data throughput. This will significantly contribute to the generalizability and scalability of DRL-based mobile robot systems. As these contributions are primarily development and implementation-focused, interested readers are encouraged to consult our open-sourced codebase\footnotemark[1] for further details.

\subsection{Transqer}

\begin{figure}
	\centering
	\includegraphics[width=0.48\textwidth]{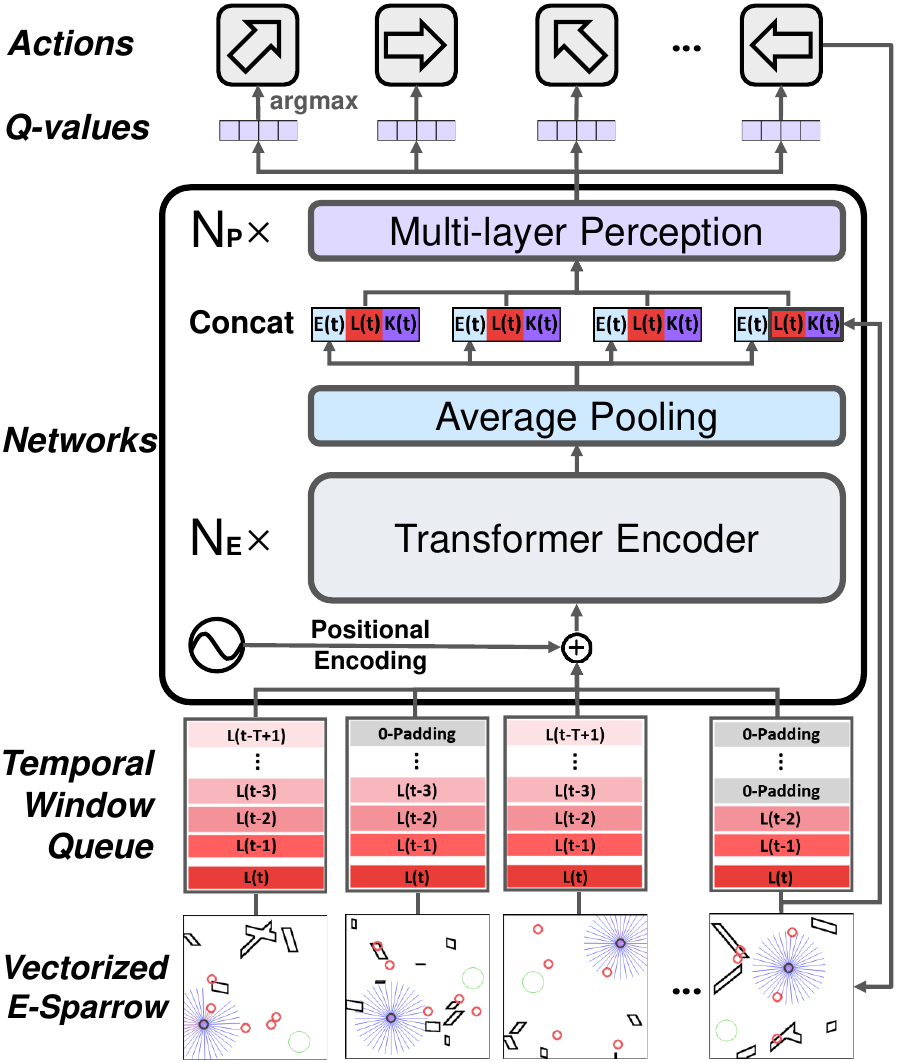}
	\caption{Architecture of the Transqer.}
	\label{transqer}
\end{figure}

This section details our proposed model, which facilitates end-to-end mapping from raw temporal perception to low-level actions. The model architecture, illustrated in Fig. \ref{transqer}, comprises three primary components: a Temporal Window Queue (TWQ), a Transformer encoder\cite{transformer}, and a Multi-Layer Perceptron (MLP) network. The TWQ, a first-in-first-out queue of length $T$, stores the temporal lidar data. The Transformer encoder, configured with $N_E$ layers, encodes this temporal information. Subsequently, the MLP network, possessing $N_P$ layers, predicts the Q-values, thereby enabling the derivation of corresponding actions.  To ensure temporal consistency, the TWQ is padded with zeros following each environmental termination.

The proposed model is named as Transqer, signifying its function of transforming temporal data into Q-values utilizing a Transformer encoder (Trans-Q-er).  During each environmental interaction, the Transqer operates as follows:

\begin{enumerate}[label=\circled{\arabic*}]
\item Acquire the most recent lidar data $L(t)$ from E-Sparrow and append it to the TWQ.
\item Integrate temporal order into the TWQ using positional encoding\cite{transformer}.
\item Encode the temporally-encoded lidar data using the Transformer encoder.
\item Aggregate the encoder output via average pooling across the temporal channel, resulting in the aggregated lidar data, $E(t)$.
\item Concatenate $E(t)$ with the current lidar data $L(t)$ and robot kinematic information $K(t)$.
\item Map the concatenated vector to Q-values using the MLP network.
\item Select actions corresponding to the maximum Q-values and execute them within the vectorized E-Sparrow environment.
\end{enumerate}

Leveraging the vectorized simulation capabilities of E-Sparrow and the parallel processing capacity of neural networks, the above procedures can be executed concurrently across multiple instances during simulation training, significantly enhancing the efficiency of the learning system. Prior to advancing, three specifics warrant further discussion.

\subsubsection{Why TWQ}
\label{WhyTWQ}
It is widely recognized that off-policy DRL algorithms demonstrate superior sample efficiency compared to on-policy methods, a feature that is essential for the practical deployment of DRL in real-world scenarios. As outlined in Section \ref{TPinDRL}, while RNN-based methods are fundamentally incompatible with off-policy algorithms and CNN-based approaches exhibit limitations in temporal modeling, this study adopts the framework of Transformer-based architectures combined with off-policy DRL. Although prior research, such as DTQN\cite{DTQN}, has demonstrated the feasibility of learning directly from consecutive samples (a continuous trajectory) with Transformer in an off-policy manner, it is inefficient due to the strong correlations inherent in consecutive samples.
Specifically, even minor modifications to the Q-value can lead to substantial shifts in the policy, thereby altering the data distribution and disrupting the relationship between the Q-value and its target\cite{DQN}. This phenomenon can lead to suboptimal policies or even unstable learning performance, which will be empirically demonstrated in our experimental section (see Fig. \ref{DTQN_Compare}).
To mitigate this challenge, we introduce a TWQ to break the continuous trajectory into independent temporal windows for each timestep $t$, as denoted by $[L(t),L(t-1),...,L(t-T+1)]$. It is important to note that this entire temporal window is used exclusively for Q-value prediction at timestep $t$, and not for predictions from timestep $t$ to $t-T+1$. Concurrently, the first-in-first-out characteristic of the TWQ guarantees the allocation of a corresponding temporal window to each timestep, thereby ensuring the training for all timestep.
These independent temporal windows are subsequently stored in a replay buffer and randomly sampled during training. This strategy circumvents the inefficiencies of learning consecutively from long sequences and simultaneously enhances the data utilization rate, ultimately bolstering performance.

\subsubsection{Why Average Pooling}
The core function of the temporal perception module (the TWQ and the Transformer encoder) is to comprehend and forecast the environmental dynamics, the result of which is a tensor of shape $(N, T, D)$, where $N$ represents the number of vectorized environments, $T$ denotes the temporal window length, and $D$ signifies the feature dimension (corresponding to the number of lidar beams in this study). Prior to Q-value prediction, this tensor must be aggregated across the temporal channel, resulting in an output tensor of shape $(N, D)$ for the MLP's inference. Three primary aggregation methods were considered: (i) average pooling, (ii) max pooling, and (iii) positional pooling (only using the most recent temporal position). Empirical results indicate that average pooling surpasses the other two methods in performance. We posit that this superiority stems from average pooling's ability to consolidate maximum information across the temporal channel. This consolidation is crucial for temporal understanding, whereas the other approaches predominantly focus on a singular, critical time point.

\subsubsection{Why concatenating}
A potential question arises regarding the necessity of concatenating $L(t)$ prior to Q-value prediction, given that $E(t)$ already encapsulates lidar information. Our rationale is that $E(t)$ primarily serves to facilitate comprehension and prediction of environmental dynamics, thereby guiding the agent's subsequent high-level decisions, such as bypassing, overtaking, or retreating. However, it may overlook crucial information for collision avoidance, specifically the instantaneous distance to obstacles. Consequently, $L(t)$ is concatenated to emphasize this information.  An ablation study was conducted in the pre-experiment to validate the hypothesis, revealing that the omission of $L(t)$ leads to a degradation in performance. Furthermore, the robot's kinematic information, $K(t)$, is also concatenated to furnish vital information for robot control and target reaching.

\subsection{Symmetric Invariance}

\begin{figure}
	\centering
	\includegraphics[width=0.48\textwidth]{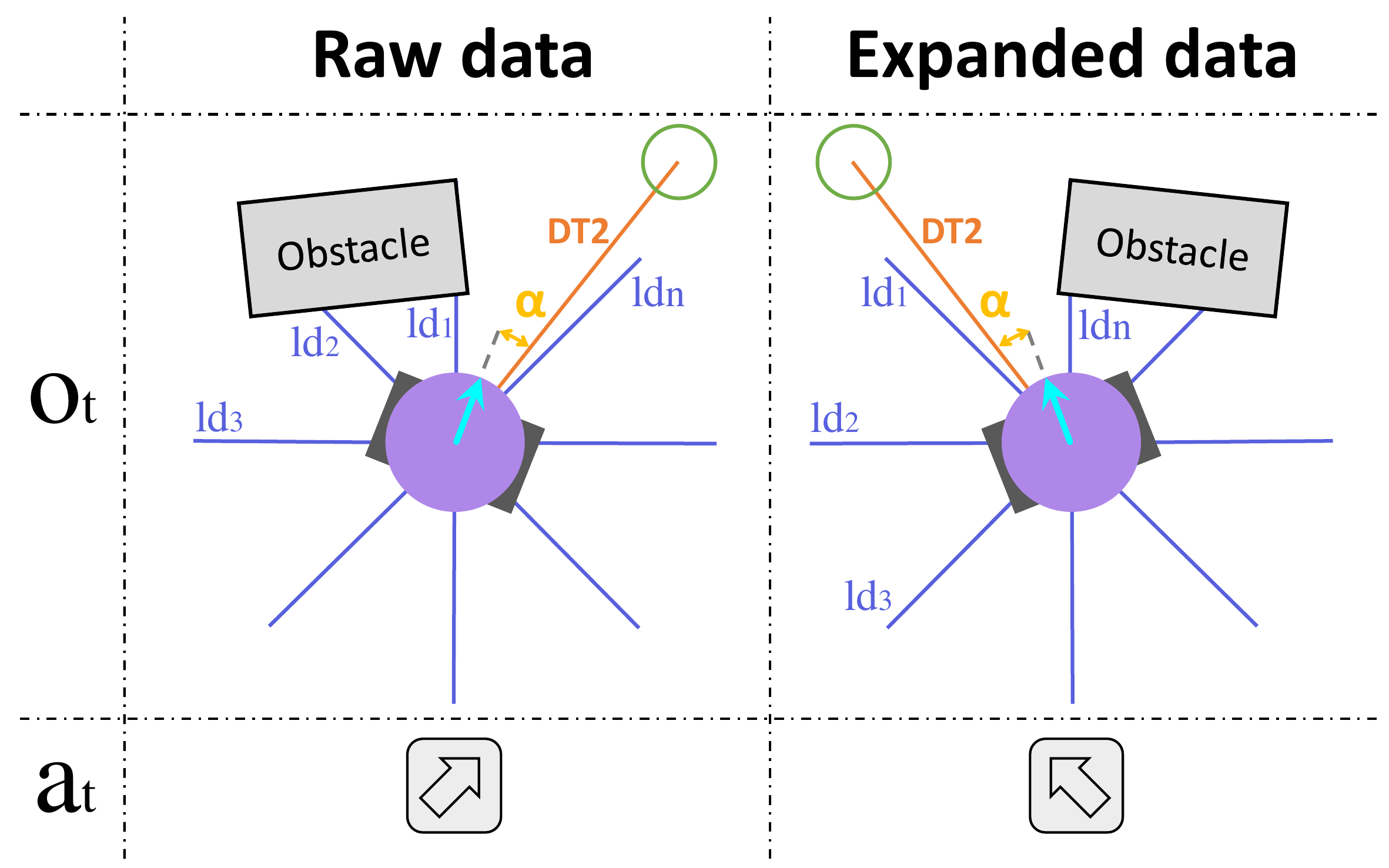}
	\caption{Symmetric Invariance.}
	\label{SI}
\end{figure}

Prior research\cite{scalinglaw,trans_survey1,trans_survey2} has indicated that Transformer models are demanding regarding training samples, both in terms of data volume and diversity. This demand can be further exacerbated within the context of online DRL due to the non-stationary nature of the training process, wherein the target Q-values associated with a given observation vary during training\cite{DQN}. To accommodate this issue, we introduce a data expansion technique predicated on Symmetric Invariance (SI), as illustrated in \mbox{Fig. \ref{SI}}. Given a raw transition $(o_t,a_t,r_t,o_{t+1})$ generated by the simulation environments, the SI-expanded transitions is given by $(o^{SI}_t,a^{SI}_t,r_t,o^{SI}_{t+1})$, where

\begin{equation}
	\label{Ot_SI}
	o^{SI}_t = [K^{SI}(t),L^{SI}(t),L^{SI}(t-1),...,L^{SI}(t-T+1)]
\end{equation}

\begin{equation}
	\label{Kt_SI}
	K^{SI}(t)=[V_l^{ct}, -V_a^{ct}, V_l^{rt}, -V_a^{rt}, D2T, -\alpha, V_l, -V_a]
\end{equation}

\begin{equation}
	\label{Lt_SI}
	L^{SI}(t) = [ld_n, \cdots, ld_3, ld_2, ld_1]
\end{equation}

\begin{equation}
	\label{At_SI}
	a^{SI}_t = [V_l^{ct}, -V_a^{ct}]
\end{equation}

In contrast to existing symmetry-based data expansion techniques\cite{symmetry1,symmetry2,symmetry3} that primarily focus on image-type perception data for supervised learning, our work extends this concept to DRL-based robotic systems and factor not only perception data but also the kinematic state of the robot.

\subsection{ColorDynamic}
\label{CD}

Integrating the end-to-end problem formulation, the E-Sparrow simulator, the Transqer network, and the SI data expansion technique, we formally present our DRL solution to the LPP, as delineated in Algorithm 1. We name our approach as ColorDynamic, where ``Color'' originates from our previous work\cite{Color} and ``Dynamic'' highlights the extension to dynamic environments.

ColorDynamic comprises two distinct phases: interaction and training. During the interaction phase, the Transqer interacts with the E-Sparrow simulation platform, aided by the TWQ, and the resultant data is stored in a replay buffer for subsequent training. In the training phase, a batch of data is sampled from the replay buffer and subsequently extended via SI according to Eq. (\ref{Ot_SI})$\sim$(\ref{At_SI}). Following this, the raw and extended data are concatenated for DRL training. For the training algorithm, we employ DQN\cite{DQN}, an off-policy DRL algorithm that has demonstrated human-level decision-making capabilities in discrete action spaces. Additionally, the Double Q-learning\cite{DDQN} technique is incorporated to mitigate the notorious overestimation issue inherent in DQN. Moreover, the Actor-Sharer-Learner training framework\cite{Color}, featuring vectorized interaction and decoupled training, is harnessed to improve training efficiency.

ColorDynamic exhibits several advantageous features: 
\mbox{\textbf{(a) Generalizability:}} The diversity provided by the E-Sparrow simulator, encompassing varied environmental layouts, sensor noise, and robot kinematics, significantly contributes to the generalizability of the trained local planner to unseen environments and to unfamiliar robot kinematics. Furthermore, this diversity can be augmented through the SI technique, thereby further improving the model's generalizability.
\textbf{(b) Scalability:} The inherent parallel execution capability of Transqer (or more broadly, neural networks) facilitates seamless extension from single-robot to multi-robot planning. 
\textbf{(c) Real-time Performance:} The end-to-end design eliminates pipeline preprocessing of raw sensor data, thereby tremendously improving real-time decision-making performance.
\textbf{(d) Dynamic Environments Adaptability:} The temporal perception module of Transqer, comprising the TWQ and Transformer encoder, equips the trained planner with a robust capacity to comprehend, predict, and make decisions within dynamic environments.
\textbf{(e) Unstructured Environments Applicability:} The planner's perception of its surroundings is accomplished through its onboard lidar instantaneously, negating the requirement for pre-constructed environmental maps or a priori knowledge of obstacles, thus enabling its applicability to unstructured environments.

\begin{algorithm}[t]
\small 
\caption{ColorDynamic}\label{algo:ColorDynamic}
\tcp{Interaction:}
$K(t), L(t)$ = \textbf{E-Sparrow}.reset()

\textbf{TWQ}.padding(0)

\textit{done} = \textbf{False}

\While{not done}{

\textbf{TWQ}.append($L(t)$)

$[L(t), L(t-1) , \cdots, L(t-T+1)]$=\textbf{TWQ}.get()

$o_t$ = $[K(t), L(t), L(t-1), \cdots, L(t-T+1)]$

$a_t = \pi(o_t; \theta)$ \tcp{Select $a_t$ with \textbf{Transqer}}

$K(t+1), L(t+1)$, $r_t$, \textit{done} = \textbf{E-Sparrow}.step($a_t$)

\textbf{ReplayBuffer}.add($o_t, a_t, r_t$, \textit{done})

$K(t), L(t)$ = $K(t+1), L(t+1)$
}

\vspace{0.4cm}
\tcp{Training:}
($o_t, a_t, r_t$, \textit{done}, $o_{t+1}$) = \textbf{ReplayBuffer}.sample()

($o^{SI}_t,a^{SI}_t,r_t$, \textit{done},$o^{SI}_{t+1}$) = \textbf{SI}($o_t, a_t, r_t$, \textit{done}, $o_{t+1}$) 

$\theta$ = \textbf{DRL}.train(($o_t, a_t, r_t$, \textit{done}, $o_{t+1}$), \\ \hspace{2cm}($o^{SI}_t,a^{SI}_t,r_t$, \textit{done},$o^{SI}_{t+1}$))
\end{algorithm}

The experimental section will validate above desirable features. Prior to this, a comparative analysis with existing relevant works, such as Color\cite{Color}, DTQN\cite{DTQN}, and DRL-VO\cite{drlvo}, is presented here to elucidate the distinctions of our work.
Specifically, ColorDynamic is an extension of our previous work, Color. While Color addressed the LPP in static environments, ColorDynamic extends this to dynamic environments. The extension is achieved through the incorporation of a new end-to-end formulation for LPP, an enhanced simulator E-Sparrow, a novel temporal model Transqer, and an effective data augmentation technique SI.
In contrast to DTQN, the primary difference lies in the temporal perception module. ColorDynamic employs a TWQ to segment consecutive temporal data into independent temporal windows, thereby avoiding ineffective learning directly from sequential data within the context of off-policy DRL. In addition, while DTQN is presented as a DRL algorithm, it lacks a detailed discussion of its practical application for LPP.
Compared to DRL-VO, ColorDynamic adopts an end-to-end architecture, contrasting with DRL-VO's reliance on an expertly-designed, complex sensor preprocessing pipeline. Additionally, ColorDynamic utilizes a computationally efficient simulator to promote training. These features significantly reduce computational demands, allowing it to be trained on a desktop-class computer rather than a workstation with multiple GPUs.  Most importantly, DRL-VO operates under the assumption that dynamic obstacles (pedestrians) will actively avoid the controlled robot. This assumption is highly problematic, as pedestrians may fail to see the robot due to inattention, compromising the safety of the resulting local planner. Conversely, ColorDynamic's training involves randomly moving dynamic obstacles. While this introduces significant challenges to the training process, as obstacles may not avoid and might even attack the agent, it fosters the learning of safer behaviors, such as active circumnavigation, overtaking, or strategic retreat.

\begin{figure}
	\centering
	\includegraphics[width=0.45\textwidth]{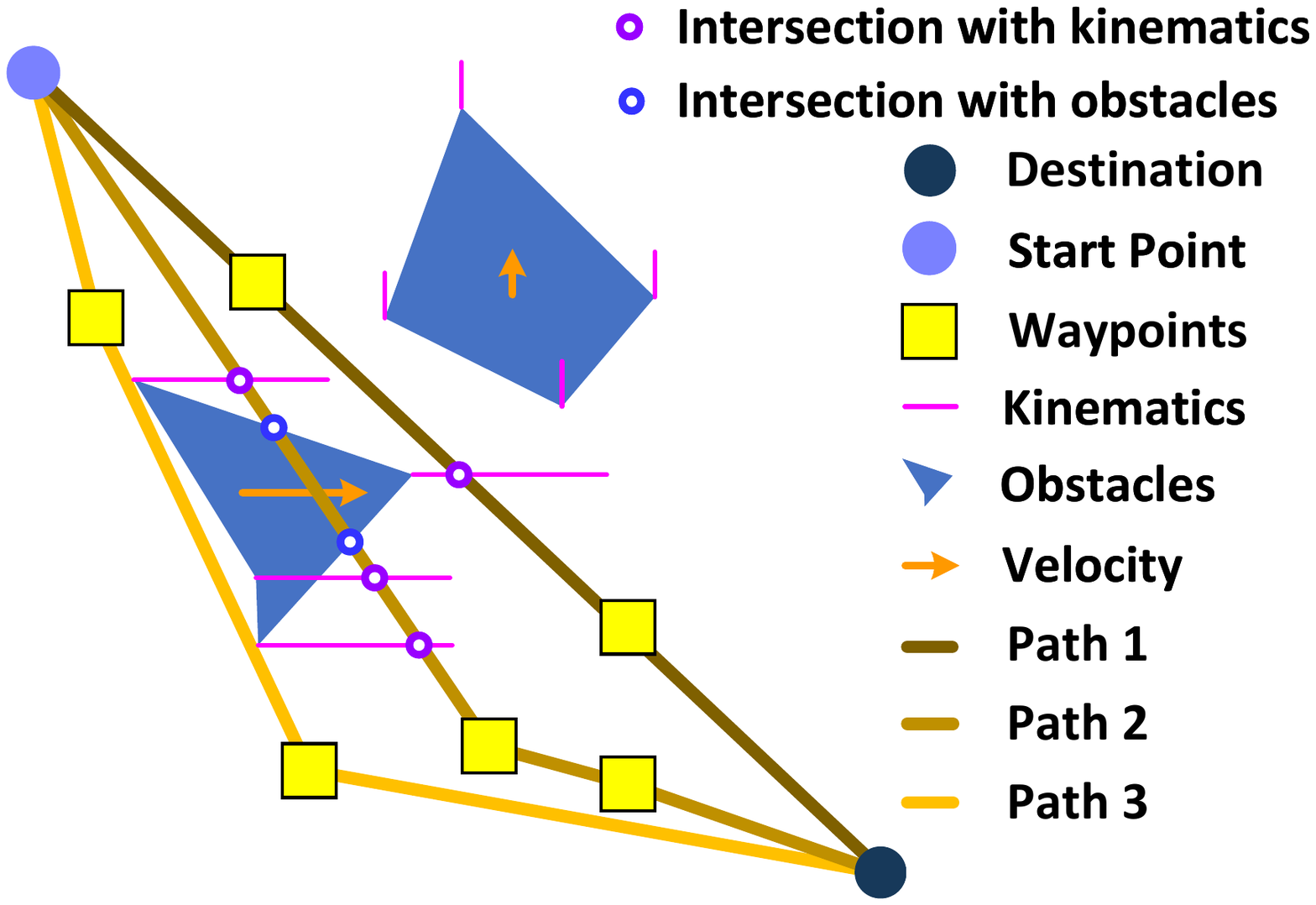}
	\caption{Diagram of OkayPlan.}
	\label{OKA}
\end{figure}

\begin{figure}
	\centering
	\includegraphics[width=0.45\textwidth]{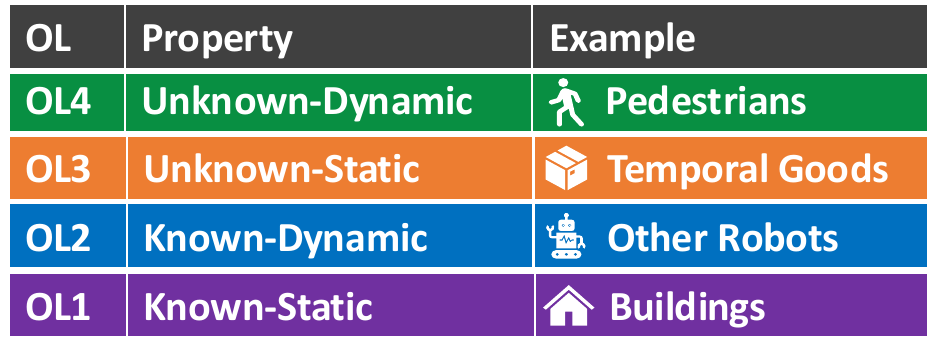}
	\caption{Obstacle layers.}
	\label{OL1234}
\end{figure}

\subsection{OPCD Navigation System}
This section introduces the OkayPlan-ColorDynamic (OPCD) navigation system, based on the proposed ColorDynamic and our previous work, OkayPlan\cite{okayplan}. In this system, OkayPlan functions as a global planner, while ColorDynamic serves as a local planner. 

\subsubsection{A Review of OkayPlan} The diagram of the OkayPlan algorithm is illustrated in Fig. \ref{OKA}, wherein the global path planning problem is formulated as an optimization problem, as follows:
\begin{equation}
	\label{OKAOP}
	F(Z) = G(Z) + \delta \cdot Q(Z)^{\sigma}+ \mu \cdot P(Z)^{\nu}
\end{equation}

\begin{figure}[t]
	\centering
	\includegraphics[width=0.45\textwidth]{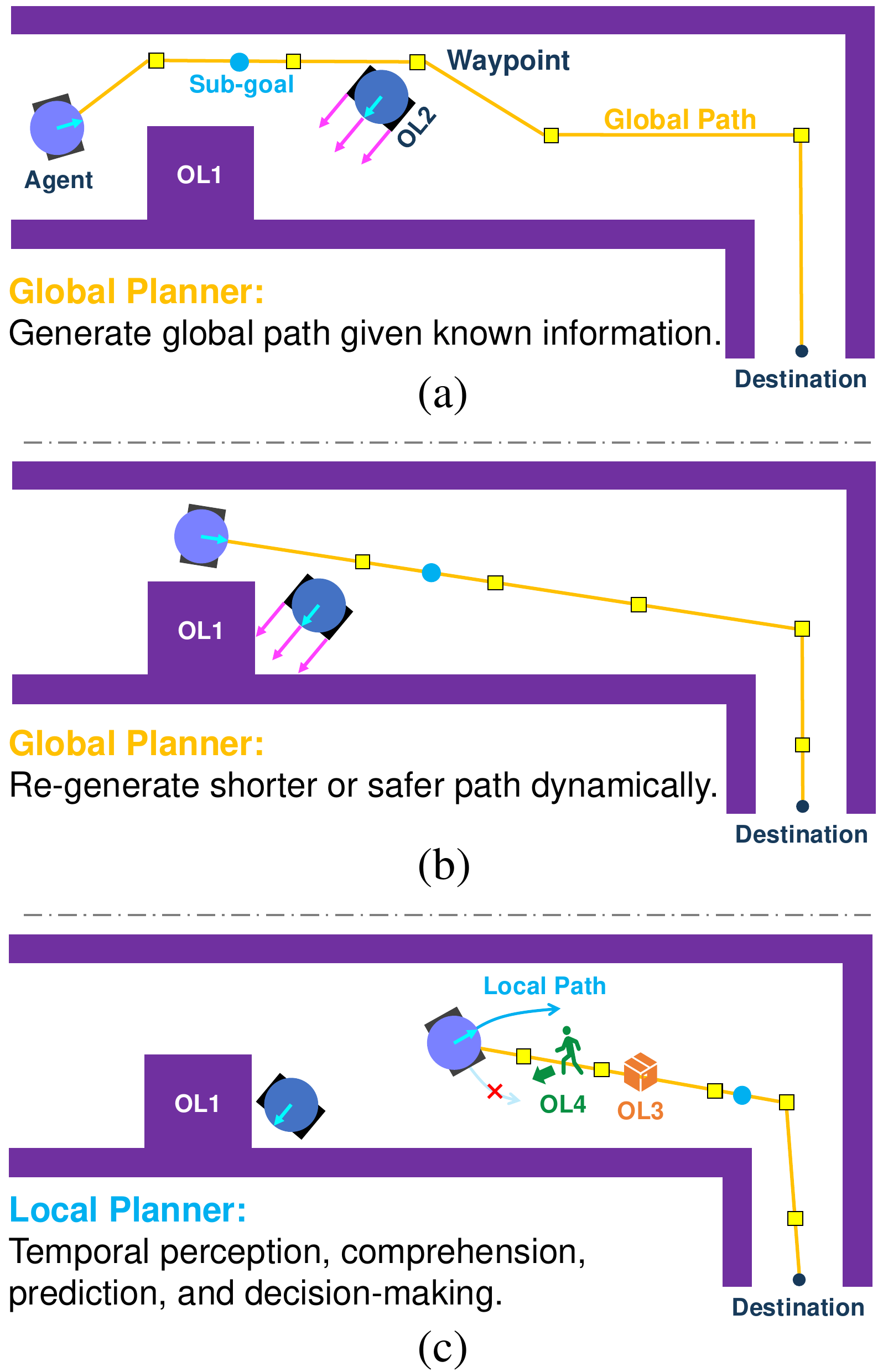}
	\caption{Illustration of the OPCD navigation system.}
	\label{OPCD}
\end{figure}

Here, $F(Z)$ represents the fitness function, with $Z$ being the decision variables representing waypoint coordinates. $G(Z)$ represents the length of path, $Q(Z)$ represents the number of intersections with obstacles, $P(Z)$ represents the number of intersections with obstacle kinematics (velocity vectors), and $\delta,\sigma,\mu,\nu$ are four coefficients. 
The optimization problem, denoted by Eq. (\ref{OKAOP}), is resolved by an efficient heuristic optimizer, SEPSO\cite{sepso}.
As exemplified by Fig. \ref{OKA}, this framework enables OkayPlan to identify a safer path \mbox{(Path 3)} in dynamic environments, as opposed to the shortest but risky path \mbox{(Path 1)}. Furthermore, the calculations of $G(Z)$, $Q(Z)$, and $P(Z)$ can be efficiently performed through matrix operations involving the waypoints, obstacle vertices, and kinematic segment endpoints, which facilitates deployment on GPUs to achieve real-time performance. It has been reported that OkayPlan is capable of generating safe yet short paths in dynamic environments at a real-time execution speed of \mbox{125 Hz}.

\subsubsection{Obstacle Definition} 
Based on the characteristics of obstacles, they are categorized into four Obstacle Layers (OL), as illustrated in Fig. \ref{OL1234}. The distinction between ``known'' and ``unknown'' properties is determined by the extent to which the navigation system has prior knowledge of the obstacles. For instance, static buildings (OL1) can be represented on a map, whereas other robotic obstacles (OL2) can communicate directly with the controlled robot. As a result, planning in known (structured) environments can be managed by the global planner. However, the global planner is only capable of generating a global path without providing control commands. Furthermore, the global path does not account for unknown obstacles. Therefore, a local planner is required for path following and collision avoidance with newly encountered unknown obstacles under unstructured environments.

\subsubsection{Integration of OkayPlan and ColorDynamic} 
Based on the aforementioned principles, the navigation system is designed and depicted in Fig. \ref{OPCD}. This configuration necessitates a global planner capable of generating a path based on known information and dynamically recalculating shorter and safer paths as the known environmental information changes, as shown in Fig. \ref{OPCD} (a) and (b). Simultaneously, it requires a local planner with real-time temporal perception capabilities to comprehend, predict, and make decisions when encountering newly emerged unknown obstacles, thereby avoiding potentially unsafe local paths, as highlighted in Fig. \ref{OPCD} (c). 
The dynamic regeneration ability is guaranteed by OkayPlan, while the real-time temporal perception capability is provided by ColorDynamic. Their integration is achieved by selecting a sub-goal along the global path at a distance of $0.6d$ from the robot, which serves as the target point for ColorDynamic. Here, $d$ is the maximum local planning distance. Recall that in the reward function, exceeding $d$ is penalized and regarded as termination. Consequently, for scenarios where $D2T > d$, the agent lacks familiarity. To account for this, a discount factor of 0.6 is applied when navigating, providing a buffer distance to accommodate potential retreats.

\begin{figure*}[t]
	\centering
	\subfloat[]{\includegraphics[width=0.33\textwidth]{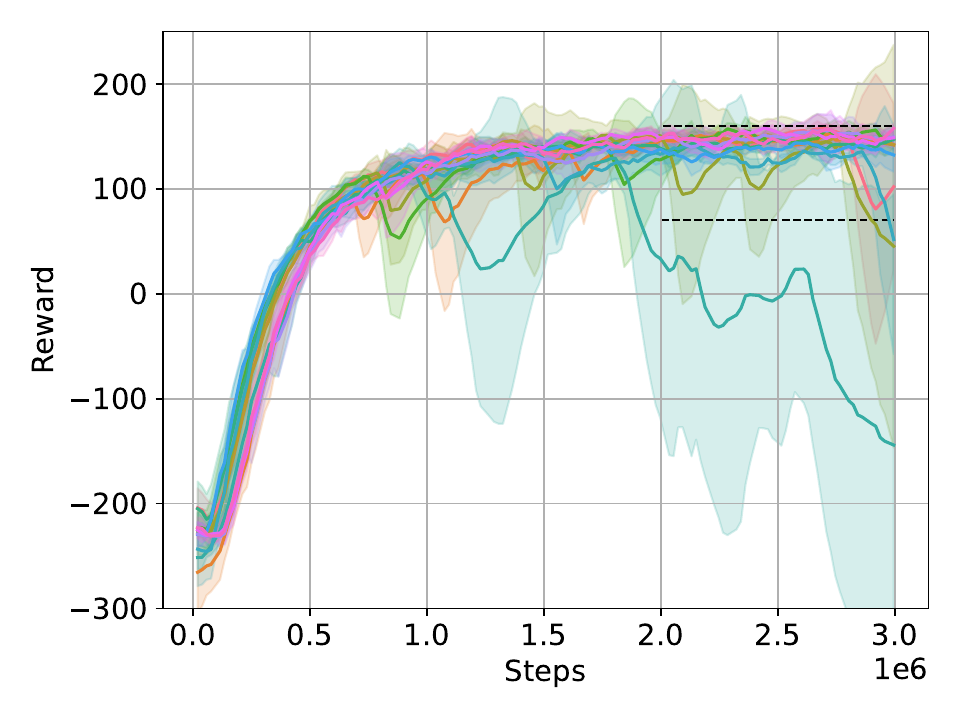}}
	\hfil
	\subfloat[]{\includegraphics[width=0.33\textwidth]{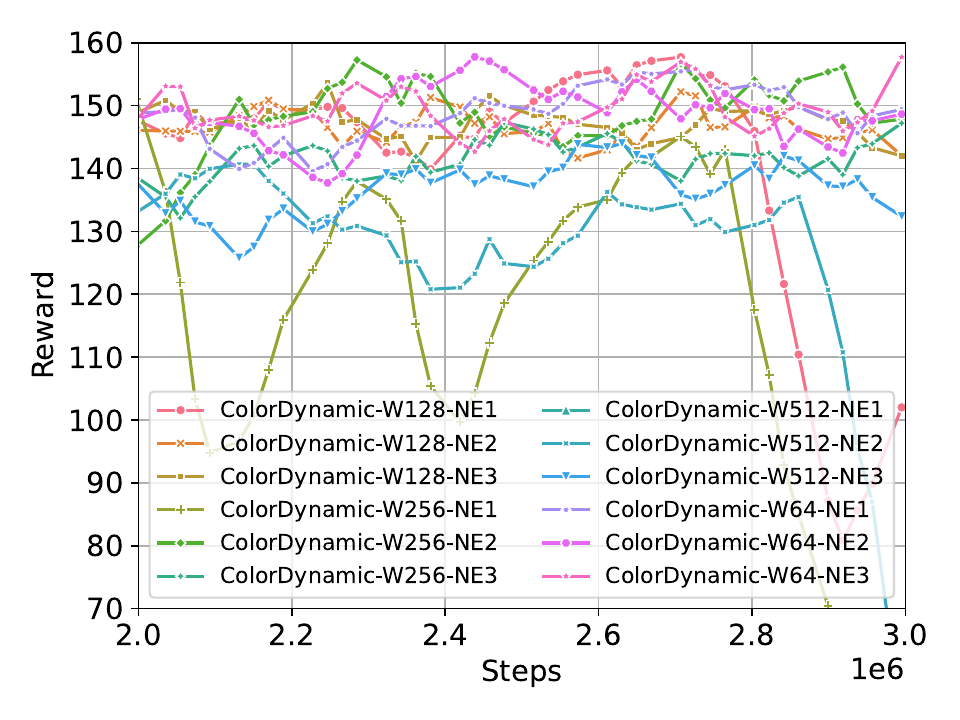}}
	\hfil
	\subfloat[]{\includegraphics[width=0.33\textwidth]{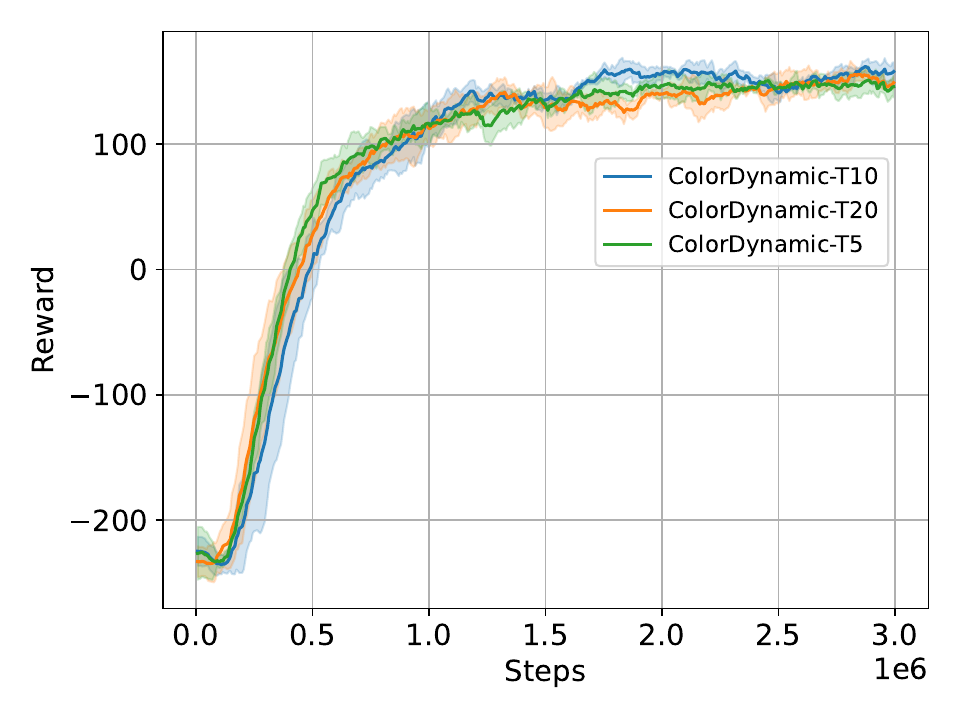}}
	\hfil
	\caption{Determination of $W$, $N_E$, and $T$. Here, (b) is a scaled snapshot of the black box in (a).}
	\label{W_NE_T}
\end{figure*}

One might argue that, since ColorDynamic demonstrates proficiency in managing unknown obstacles, it should logically be able to handle known obstacles as well. If this assumption holds, then what is the necessity of OkayPlan? Being a local planner, in order to achieve real-time performance in unstructured environments, ColorDynamic is not integrated with a global map. Consequently, it is limited by a maximum local planning distance $d$, which can pose challenges for long-distance planning within OL1. Note that a greedy increase in $d$ during training can lead to a degradation in performance. Concerning OL2, ColorDynamic is capable of addressing them independently. However, the OkayPlan from a higher and more global level can assist the local planner in avoiding high-risk collision areas proactively in advance, as will be validated in the experiments (see Fig. \ref{dangerous_zoom}).

\section{Local Planning with ColorDynamic}
This section assesses the proposed ColorDynamic, focusing on the determination of hyperparameters, the evaluation of the trained local planner, the comparison with existing approaches, and the ablation of the proposed techniques. The software and hardware configurations supporting these experiments are summarized in Table \ref{tab:hard_software} in the Appendix.

\subsection{Experimental Setup}
\label{ExperimentalSetup}
The proposed E-Sparrow simulator is employed to train and evaluate ColorDynamic in the context of local planning, with the experimental setups detailed as follows.

\subsubsection{Environment}
The training map is configured with a size of 8 $m$, and the maximum local planning distance $d$ is set to 4 $m$. 
The number of static obstacles (OL3) is randomly sampled between 0 and 36. The number of dynamic obstacles (OL4) is set to 15, with a maximum velocity of 0.5 $m/s$, moving randomly within the map. 
Notably, the shape and location of obstacles are procedurally generated at the initialization of the environment. 
Episodes are truncated if the interaction steps exceed 500. The target is deemed reached when $D2T \leq 0.3\ m$.

\subsubsection{Robot}
\label{robot_specification}
The robot has a diameter of 0.2 $m$, with maximum linear and angular velocities constrained to 0.5 $m/s$ and 2 $rad/s$, respectively. Such configuration is derived from our real-world robot, as shown in Fig. \ref{fml_orin}(a). The control frequency is set to 10 Hz. The lidar scanning range covers $360^{\circ} \times 3m$ and consists of 24 lidar beams in total. The robot's initial pose is randomly generated at the commencement of each interaction episode. The robot is trained to repeatedly reach the target point. When reached, the target point will be randomly generated within $d$ around the robot.

\subsubsection{Hyperparameters}
The maximum training steps are set to 3M, with a learning rate of 0.0001 and a batch size of 256. During training, the environment is reset every 32K steps to regenerate obstacles randomly, and the model is evaluated every 6.4K steps. The number of vectorized simulation environments, $N$ is set to 32 for training and 10 for evaluation. Note that these hyperparameters were manually tuned. The agent is trained using the DQN algorithm, leveraging Double Q-learning and ASL techniques as introduced in Section \ref{CD}. 

\subsection{Model Complexity}
\label{ModelComplexity}
Determining an appropriate model complexity is crucial for the efficacy of ColorDynamic. The model incorporates six critical hyperparameters to define its complexity:

\begin{itemize}
\item $T$: temporal perception length.
\item $W$: width of linear layers for both the Transformer encoder and MLP.
\item $N_E$: number of Transformer encoder layers.
\item $N_P$: number of MLP layers.
\item $H$: number of heads in the multi-head attention mechanism\cite{transformer} within the Transformer encoder.
\item $D$: feature dimension of the Transformer encoder.
\end{itemize}

A comprehensive grid search across these hyperparameters is computationally prohibitive. Therefore, certain values are adopted empirically from existing studies, while the remaining are determined through a coarse search. Specifically, $N_P$ is set to 3, referencing Color\cite{Color}; $H$ is set to 8, referencing Transformer\cite{transformer}; and $D$ is simplified to match the number of lidar beams, which is 24 in this study. Regarding the $W$ and $N_E$, a grid search was conducted over the parameter space \mbox{$[64,128,256,512] \times [1,2,3]$}, as illustrated in Fig. \ref{W_NE_T}. 
Note that these curves are smoothed with an exponential factor of 0.95 for better readability. Meanwhile, these experiments were conducted across three random seeds, with the solid curves representing the mean values and the translucent shaded areas denoting the standard deviation. Same data processing procedures were applied to Fig. \ref{DTQN_Compare} and Fig. \ref{ablation} as well.
Subsequently, the temporal window length $T$ was determined through a search over $[5,10,20]$, which corresponds to temporal perception durations of $[0.5,1, 2]s$, considering the control frequency is 10 Hz. The results of this search are shown in \mbox{Fig. \ref{W_NE_T}(c)}.
The results reveal that the configuration \mbox{$[W=64, N_E=3, T=10]$} delivers the best overall performance in terms of both final reward and training stability. This configuration is therefore adopted as the default for ColorDynamic and for subsequent experiments. 

\subsection{Transqer v.s. DTQN}
\begin{figure}
	\centering
	\includegraphics[width=0.5\textwidth]{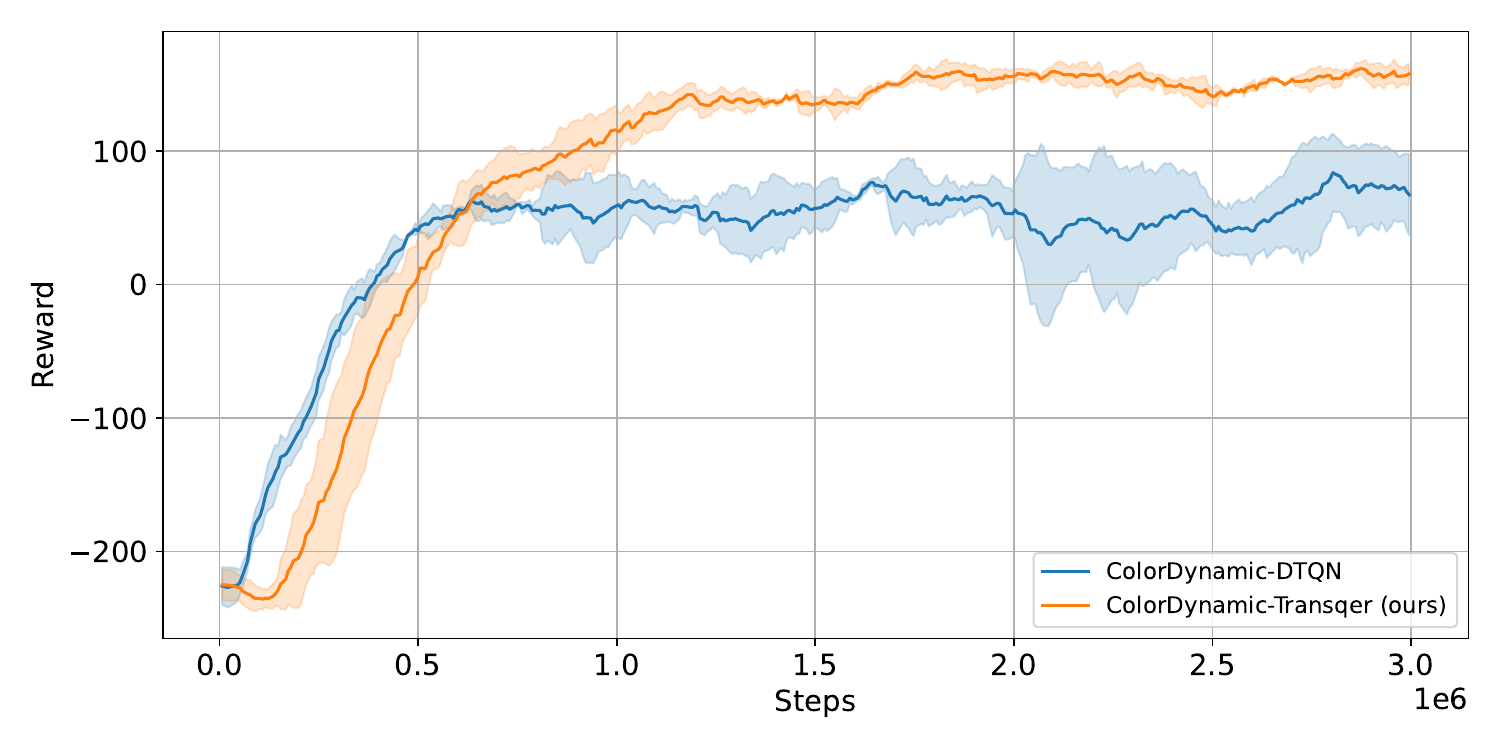}
	\caption{Comparison between Transqer and DTQN.}
	\label{DTQN_Compare}
\end{figure}

ColorDynamic harnesses Transqer as its core model. The work most similar to Transqer is DTQN\cite{DTQN}. Therefore, this section presents a comparison of Transqer and DTQN within the ColorDynamic framework in the context of LPP. The experiment adheres to the configurations outlined in Sections \ref{ExperimentalSetup} and \ref{ModelComplexity}, with results shown in Fig. \ref{DTQN_Compare}.

The results indicate a considerable enhancement in Transqer’s performance over DTQN, both in terms of final performance and training stability. We contend such improvement is attributed to Transqer’s segmentation of long temporal consequence into independent temporal windows, thereby mitigating the inefficiencies associated with direct learning from consecutive data, as discussed in Section \ref{WhyTWQ}.

\subsection{Evaluation of ColorDynamic}
\label{Evaluation_ColorDynamic}
This section assesses the local planner trained by ColorDynamic. To this end, we introduce three evaluation metrics:
\begin{itemize}
\item \textbf{Average speed}: the mean velocity of the controlled robot.
\item \textbf{Average planning time}: the mean planning time required by the planner, specifically the duration of $a_t = \pi(o_t; \theta)$.
\item \textbf{Success rate}: the success rate of the local planner in controlling the robot from a random start point to a random target point at a distance of 2 $m$. This distance is adopted from DRL-VO\cite{drlvo} for a fair comparison.
\end{itemize}

The selection of the optimal model trained by ColorDynamic is performed by evaluating the model every 6.4K environmental interactions during training. Subsequently, the best model is evaluated and compared under different configurations. To enhance reliability, the selection, evaluation, and comparison are performed over 100 random maps.

\subsubsection{Case Study}
\begin{figure*}
	\centering
	\includegraphics[width=0.95\textwidth]{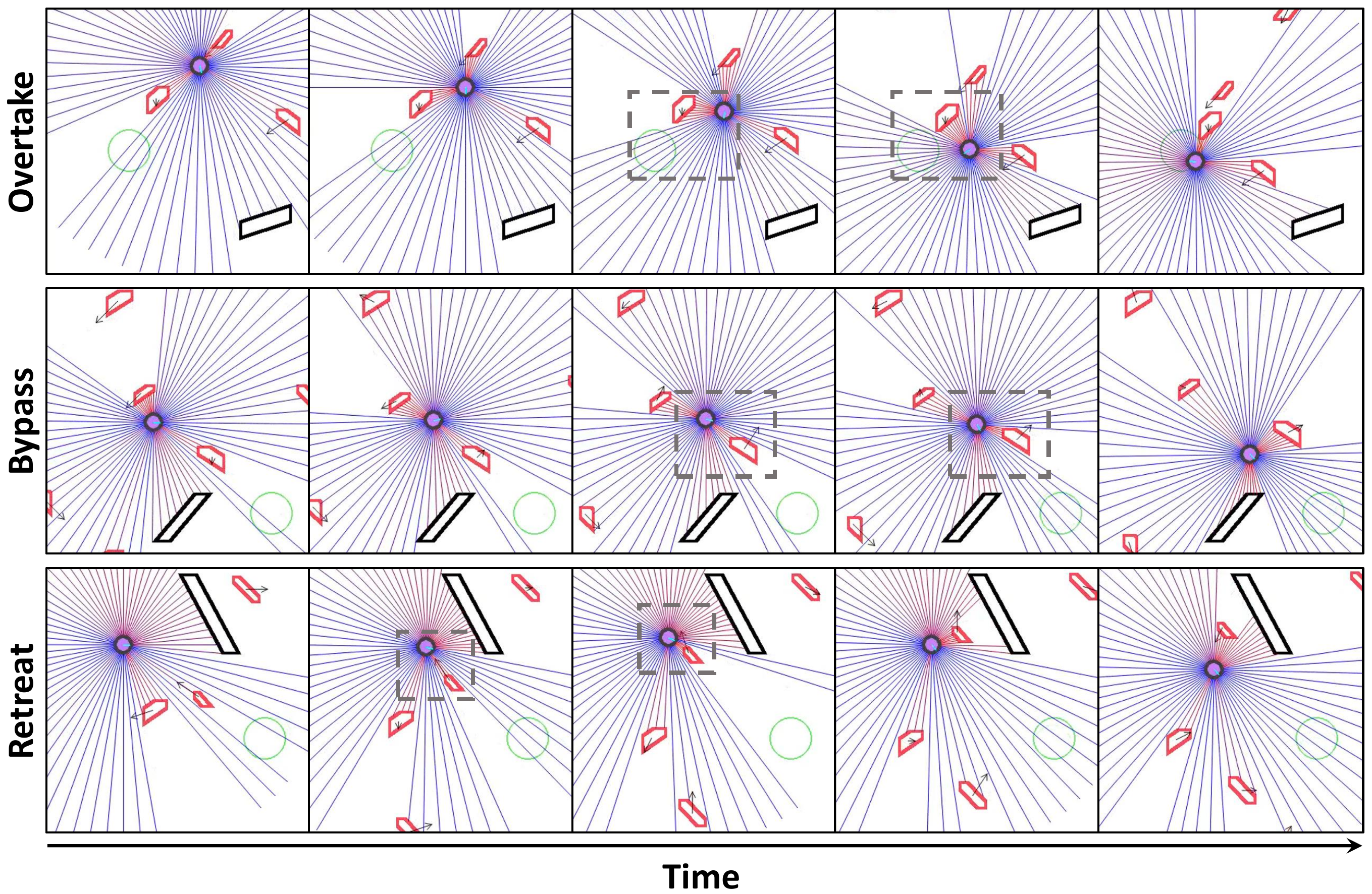}
	\caption{An exhibition of the high-level skills learned by ColorDynamic.}
	\label{skill}
\end{figure*}

In this section, we present three case studies to demonstrate the high-level skills learned by ColorDynamic. As illustrated in Fig. \ref{skill}, the model effectively \textit{overtakes} slowly moving dynamic obstacles. In scenarios where a dynamic obstacle abruptly changes direction, creating a potential collision risk, the model adapts by altering its heading to \textit{bypass} the obstacle. Additionally, when an obstacle moves directly toward the agent, the model strategically \textit{retreats} to avoid being attacked. These advanced capabilities are attributed to the model's comprehensive understanding of environmental dynamics, which substantiates the effectiveness of the temporal perception capabilities of the Transqer networks.

\begin{table*}
\centering
\caption{Comparison of Different Local Planners.}
\label{tab:lpc}
\begin{tabular}{lllll}
\hline
\textbf{Method} & \textbf{Map type} & \textbf{Average speed (m/s)} & \textbf{Average planning time (ms)} & \textbf{Success rate}\\ \hline
DWA\cite{DWA} & Fig. \ref{SparrowCompare}(c), 8m$\times$8m & 0.44 & 136.79 & 0.77 \\
APF\cite{APF} & Fig. \ref{SparrowCompare}(c), 8m$\times$8m & \textbf{0.45} & 1.52 & 0.73 \\ 
DRL-VO\cite{drlvo} & Fig. \ref{SparrowCompare}(c), 8m$\times$8m & 0.41 & 10.81 & 0.82 \\ 
Color\cite{Color} & Fig. \ref{SparrowCompare}(c), 8m$\times$8m & 0.42 & \textbf{0.24} & 0.72 \\ 
ColorDynamic & Fig. \ref{SparrowCompare}(c), 8m$\times$8m & \textbf{0.45} & 1.21 & \textbf{0.93} \\ 
\hline
\end{tabular}
\end{table*}

\begin{table*}
\centering
\caption{Evaluation Result across Different Maps.}
\label{tab:generalization}
\begin{tabular}{lllll}
\hline
\textbf{Case} & \textbf{Map type} & \textbf{Average speed (m/s)} & \textbf{Average planning time (ms)} & \textbf{Success rate}\\ \hline
1: Spacious & Fig. \ref{SparrowCompare}(b), 8m$\times$8m & 0.45 & 1.21 & 1.00 \\
2: Moderate & Fig. \ref{SparrowCompare}(c), 8m$\times$8m & 0.44 & 1.21 & 0.93 \\ 
3: Crowded & Fig. \ref{SparrowCompare}(d), 8m$\times$8m & 0.43 & 1.22 & 0.91 \\ 
4: Small & Fig. \ref{SparrowCompare}(c), 4m$\times$4m & 0.43 & 1.21 & 0.92 \\ 
5: Big & Fig. \ref{SparrowCompare}(c), 12m$\times$12m & 0.44 & 1.22 & 0.96 \\ 
\hline
\end{tabular}
\end{table*}

\begin{table*}
\centering
\caption{Evaluation Result across Different Robot Numbers.}
\label{tab:scalability}
\begin{tabular}{lllll}
\hline
\textbf{Case} & \textbf{Map type} & \textbf{Average speed (m/s)} & \textbf{Average planning time (ms)} & \textbf{Success rate}\\ \hline
6: 5 Robots & Fig. \ref{SparrowCompare}(f), 8m$\times$8m & 0.44 & 1.28 & 0.98 \\
7: 10 Robots & Fig. \ref{SparrowCompare}(f), 8m$\times$8m & 0.43 & 1.29 & 0.97 \\ 
8: 20 Robots & Fig. \ref{SparrowCompare}(f), 8m$\times$8m & 0.41 & 1.31 & 0.95 \\ 
9: 30 Robots & Fig. \ref{SparrowCompare}(f), 8m$\times$8m & 0.39 & 1.31 & 0.95 \\ 
10: 40 Robots & Fig. \ref{SparrowCompare}(f), 8m$\times$8m & 0.38 & 1.33 & 0.93 \\ 
\hline
\end{tabular}
\end{table*}

\subsubsection{Comparison Study}

To demonstrate the superiority of ColorDynamic, we compare its performance with two classical local planners, APF\cite{APF} and DWA\cite{DWA}, as well as two state-of-the-art learning-based methods, Color\cite{Color} and DRL-VO\cite{drlvo}. Note that the open-source models of Color\footnote{https://github.com/XinJingHao/Sparrow-V1/tree/Sparrow-V1.2} and DRL-VO\footnote{https://github.com/TempleRAIL/drl\_vo\_nav} are employed for comparison without further modifications. The results, summarized in Table \ref{tab:lpc}, indicate that our model outperforms all other approaches in both average speed and success rate, highlighting its efficiency and safety. The superiorities stem from the E-Sparrow simulator and the Transqer networks. The PGVD feature of E-Sparrow exposes the agent to an unlimited variety of cases, while the Transqer networks facilitate effective learning from these cases by enabling comprehension, prediction, and decision-making with temporal data.
Regarding planning speed, although ColorDynamic employs a more complex network than its predecessor Color, resulting in a marginally slower processing time, it still achieves a highly real-time performance of 1.21 $ms$ per planning. This remarkable efficiency is underpinned by the end-to-end architecture of ColorDynamic. In contrast to the pipeline architecture employed by DRL-VO, our approach achieves approximately ten times faster planning speed.

\subsubsection{Generalizability}
This section evaluates the generalizability of ColorDynamic by assessing its effectiveness across various operating conditions. Specifically, we employed five types of maps, as detailed in Table \ref{tab:generalization}. Cases 1 to 3 examine variations in obstacle shape and density, while Cases 4 and 5 focus on differences in map size. The results demonstrate that ColorDynamic maintains consistent performance in terms of average speed and planning time across diverse working conditions. Although the success rate exhibits moderate fluctuations, our approach achieves a success rate exceeding 90\% even under the most challenging crowded map scenario (Case 3).
These favorable outcomes confirm the strong generalizability of ColorDynamic. We attribute this capability primarily to the PGVD, which enhances the agent’s adaptability by providing diverse learning data derived from a wide range of scenarios.
Meanwhile, the SI technique also contributes significantly by (i) augmenting and expanding the training data and (ii) mitigating biased planning policies through the provision of symmetric learning data. This approach ensures a more balanced exploration of the environment, ultimately enhancing the model’s adaptability.

\subsubsection{Scalability}
Owing to the end-to-end design of ColorDynamic and the inherent parallel execution capabilities of neural networks, the trained model can be effortlessly scaled to multi-robot local planning scenarios. As depicted in \mbox{Fig. \ref{transqer}}, this scalability is achieved by assembling observations from multiple robots and performing calculations in a batched manner. Table \ref{tab:scalability} presents the performance of ColorDynamic across scenarios involving 5 to 40 robots. Note that, in multi-robot cases, the robots are placed in a single map and deem each other as dynamic obstacles, as illustrated in \mbox{Fig. \ref{SparrowCompare}(e)}. Remarkably, despite being trained only in single-robot scenarios, the model successfully handles multi-robot cases without noticeable performance degradation.
 Furthermore, the planning time remains broadly unaffected by the increasing number of robots, thanks to the parallel processing capability. These results establish ColorDynamic as a robust and efficient local planner for multi-robot scenarios.


\subsubsection{Real-time Performance}
The average planning time presented in Tables \ref{tab:lpc}$\sim$\ref{tab:scalability} substantiates the real-time performance of ColorDynamic, demonstrating that each planning circle can be completed within 1.2-1.3 $ms$. It should be noted that the average planning time in Table \ref{tab:scalability} represents the total planning time for all robots, rather than for a single robot. Two key features of ColorDynamic contribute to its favorable real-time performance. First, the end-to-end formulation eliminates complex sensor preprocessing procedures, thereby significantly enhancing computational efficiency. Second, ColorDynamic employs neural networks as its planning model, which are renowned for their fast inference capabilities and efficient parallel processing of multiple (batched) inputs.

\subsection{Ablation Study}

This section aims to examine the extent to which the proposed techniques contribute to ColorDynamic. To this end, we formulated the following ablation cases:
\begin{itemize}
\item \textbf{No Dynamic Obstacles}: The velocities of the dynamic obstacles are set to zero throughout training.
\item \textbf{No PGVD}: The diversification (reset) frequency of obstacles is reduced from 32K steps to 320K steps. In this case, we maintain a low level of diversity rather than completely eliminating it to ensure effective training.
\item \textbf{No SI}: The SI data expansion technique is excluded from the training process.
\item \textbf{No Transqer}: The Transqer is substituted with a simple MLP network characterized by $N_E$ layers and $W$ width.
\end{itemize}

\begin{figure}
	\centering
	\includegraphics[width=0.5\textwidth]{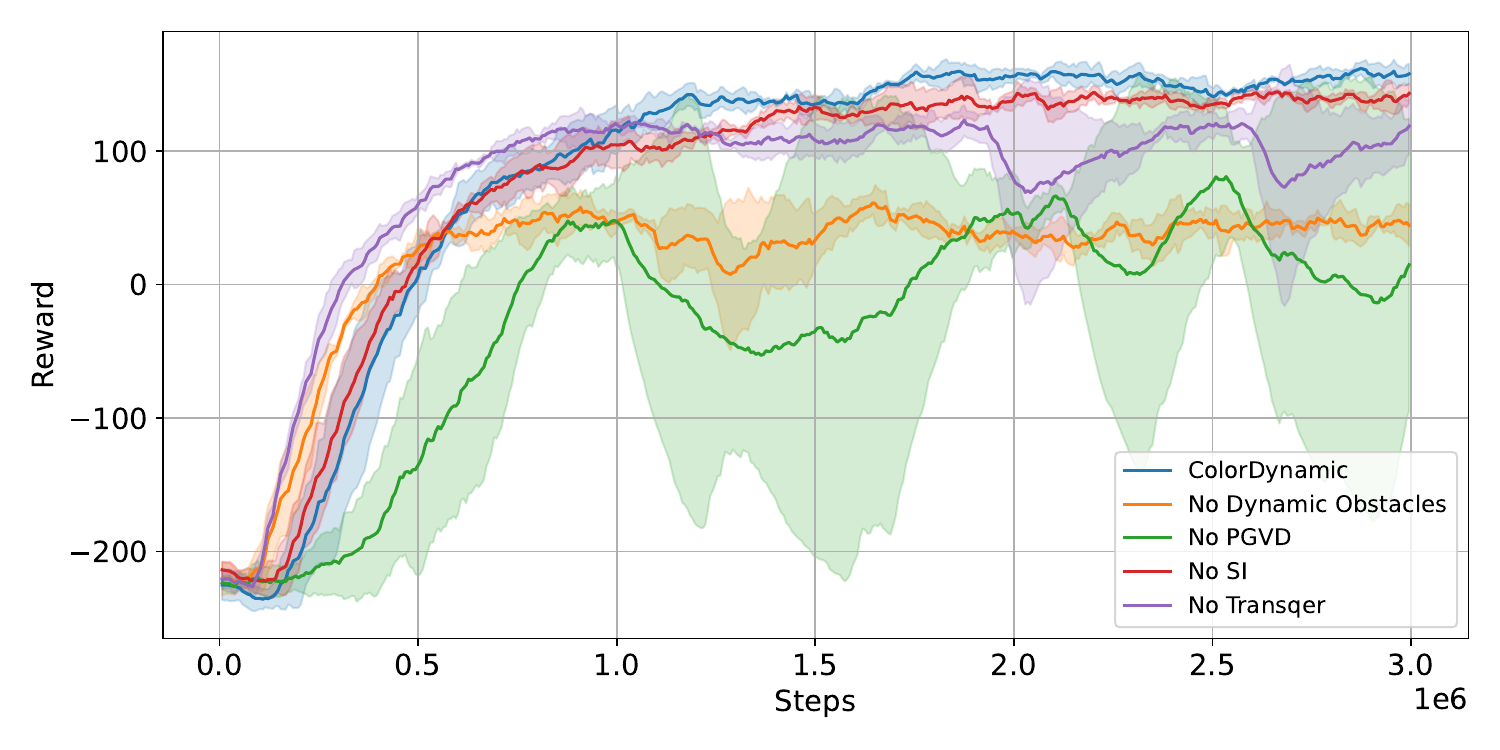}
	\caption{Ablation Results of ColorDynamic}
	\label{ablation}
\end{figure}

Note that the evaluation setups of these ablation cases are identical to those introduced in Section \ref{ExperimentalSetup}.
The ablation results illustrated in Fig. \ref{ablation} demonstrate that data quality has the most substantial impact on performance. Specifically, the removal of dynamic obstacles leads to suboptimal policies. We have observed that these policies tend to exhibit risky behaviors, such as approaching dynamic obstacles too closely. 
On the other hand, the absence of PGVD results in significant fluctuations, primarily because the trained model overfits to the training environments and fails to generalize effectively to new scenarios when obstacles are regenerated.
Fig. \ref{ablation} also indicates that the Transqer network is the third most influential component in the performance of ColorDynamic. While a simple MLP network can handle dynamic environments effectively, given its high-speed replanning characteristic, it lacks temporal perception capabilities. This deficiency impairs its ability to learn high-level skills, as depicted in Fig. \ref{skill}, ultimately degrading its performance.
Lastly, the SI technique also demonstrates efficacy, yielding consistent improvement in the later training stage.

\section{Navigating with OPCD}
\label{Sec-OPCD}

This section aims to demonstrate the superiority and applicability of the proposed OPCD navigation system. To achieve this, we initially utilize the Gazebo simulator to benchmark the OPCD against off-the-shelf navigation systems, followed by the empirical testing of OPCD in real-world scenarios. The software and hardware platforms for these experiments are reported in Table \ref{tab:hard_software} in the Appendix.

\subsection{Simulation Comparison of OPCD}

\begin{figure*}[t]
	\centering
	\subfloat[]{\includegraphics[width=0.25\textwidth]{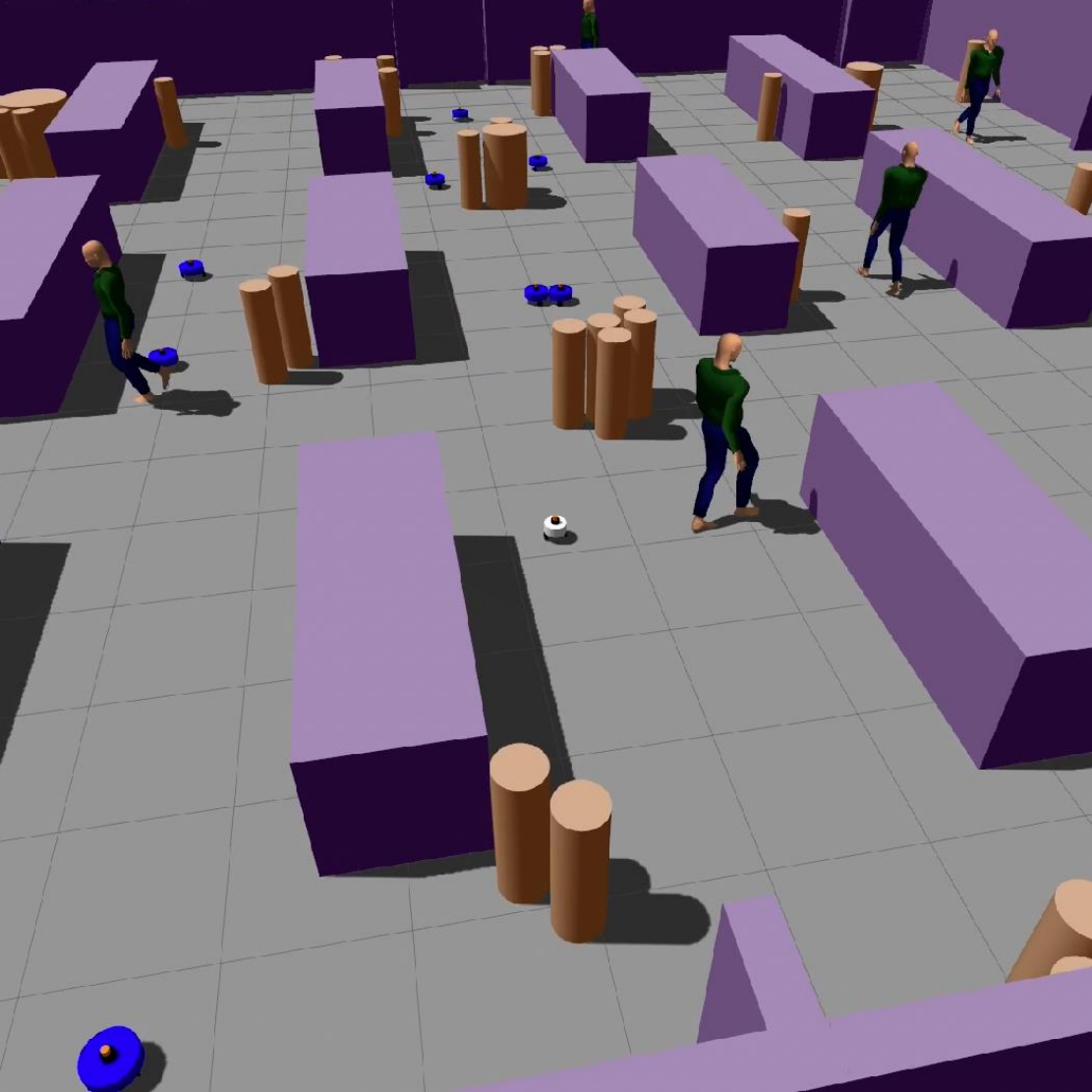}}
	\hfil
	\subfloat[]{\includegraphics[width=0.25\textwidth]{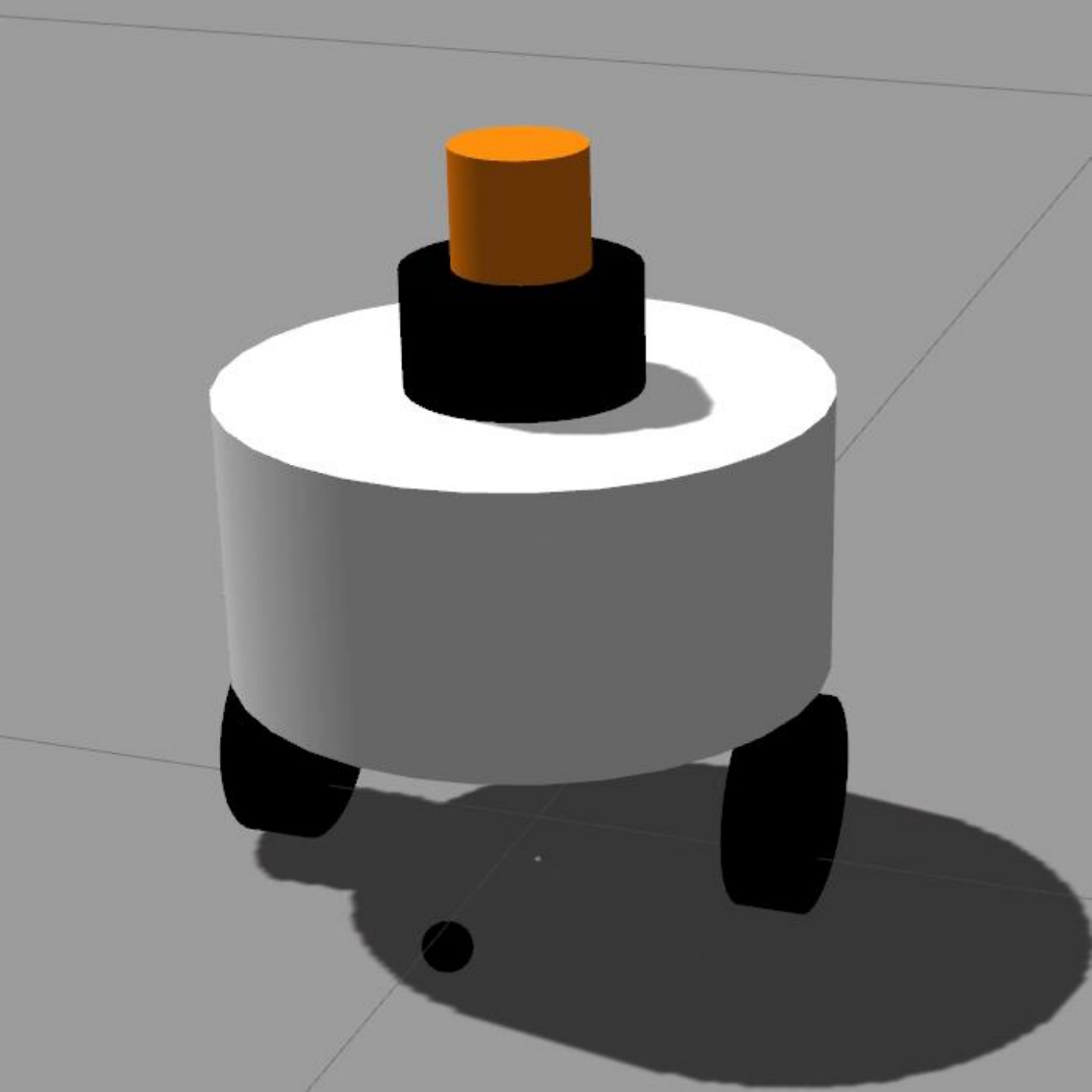}}
	\hfil
	\subfloat[]{\includegraphics[width=0.25\textwidth]{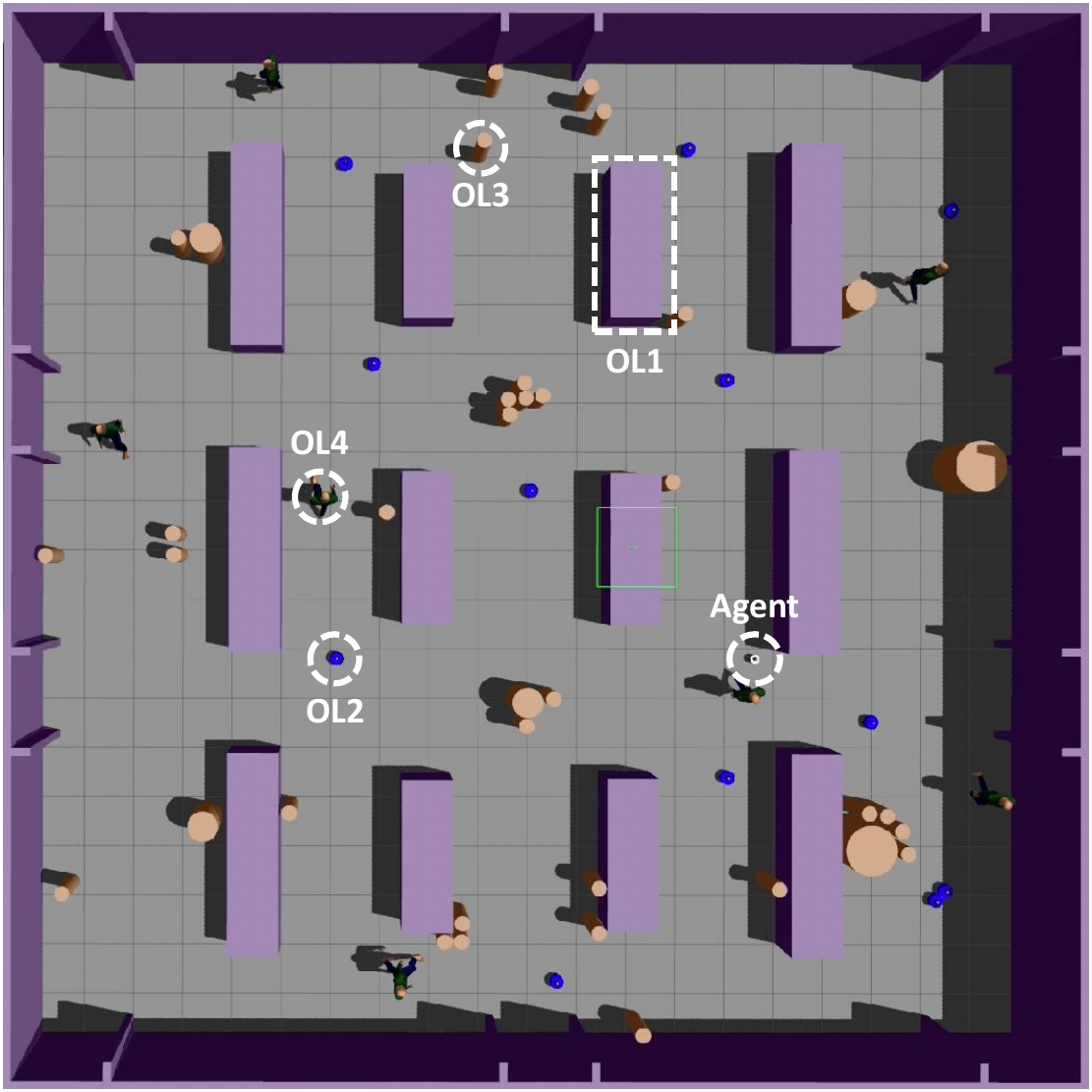}}
	\hfil
	\subfloat[]{\includegraphics[width=0.25\textwidth]{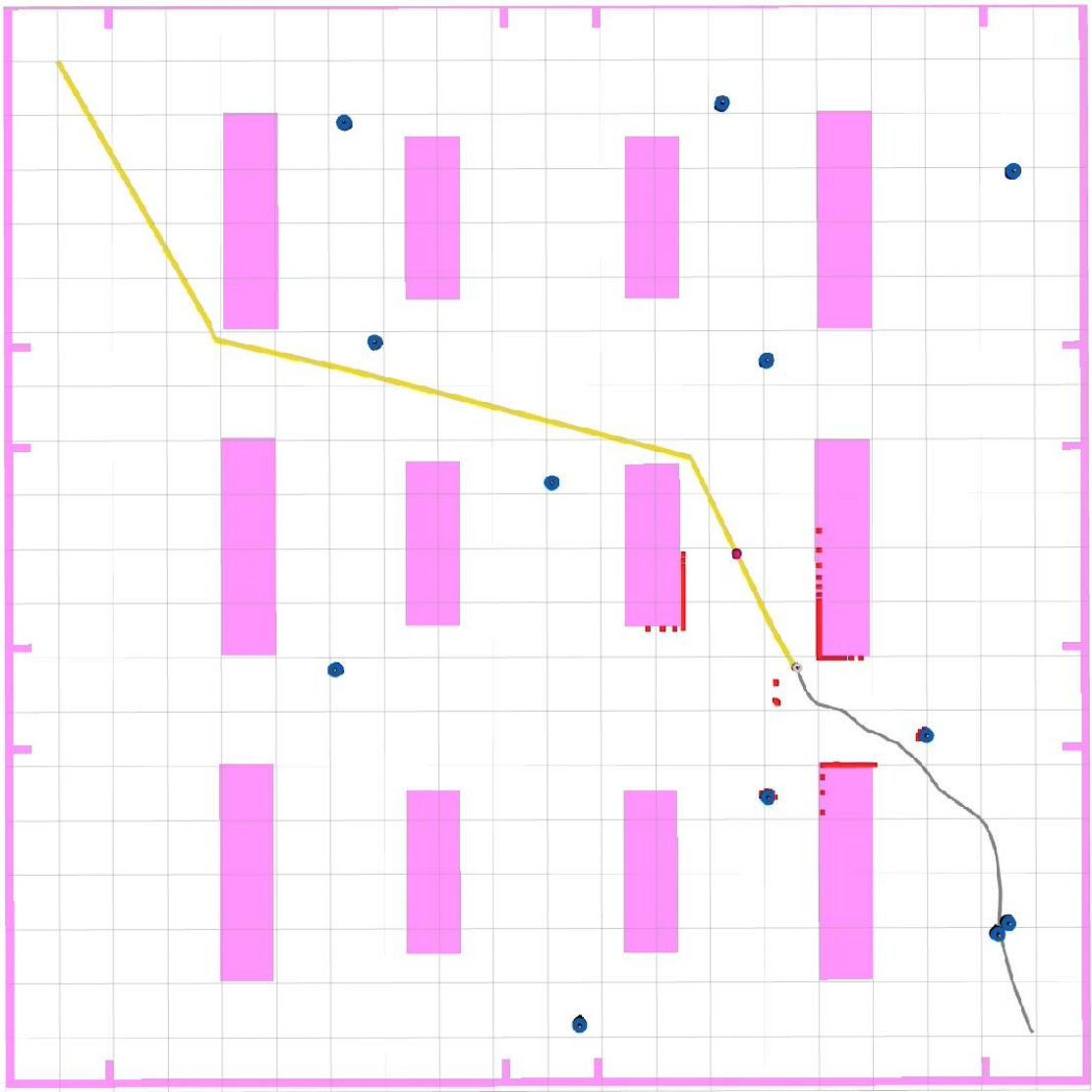}}
	\hfil
	\caption{The evaluation environment for OPCD. (a) and (b) are 3D snapshots of the Gazebo simulation environment and the two-wheeled differential robot (agent), respectively. (c) and (d) are bird’s eye views of the Gazebo simulator and Rviz visualizer. In (d), the yellow lines are the global path generated by OkayPlan, the red dot on the global path is the sub-goal for ColorDynamic, the gray lines are the historical trajectory produced by the ColorDynamic, the red grid is the lidar scanning result. Here, OL3 and OL4 are not visualized in Rviz as they are unknown to the agent.}
	\label{warehouse}
\end{figure*}

\begin{table*}
\caption{Comparison of Different Navigation Systems.}
\resizebox{\textwidth}{!}{
\begin{tabular}{lllllll}
\hline
\textbf{System} & \textbf{Global planner} & \textbf{Local planner} & \textbf{Average distance (m)} & \textbf{Average time (s)} & \textbf{Average speed (m/s)} & \textbf{Success rate} \\
\hline
1 & OkayPlan\cite{okayplan} & ColorDynamic & 30.6 & 64.1 & 0.48 & \textbf{0.96} \\
2 & OkayPlan\cite{okayplan} & Color\cite{Color} & 33.2 & 72.5 & 0.46 & 0.55 \\
3 & Voronoi\cite{voronoi} & PID\cite{pid1} & 36.0 & 77.2 & 0.47 & 0.45 \\
4 & Hybrid A*\cite{hastar} & MPC\cite{mpc} & 27.3 & \textbf{54.9} & \textbf{0.50} & 0.33 \\
5 & A*\cite{astar} & APF\cite{APF} & 28.5 & 57.8 & 0.49 & 0.26 \\
6 & Theta*\cite{thetastar} & RPP\cite{rpp} & \textbf{26.8} & 58.3 & 0.46 & 0.25 \\
7 & IRRT*\cite{irrtstar} & LQR\cite{lqr} & 29.0 & 58.3 & \textbf{0.50} & 0.10 \\
\hline
\end{tabular}}
\label{navigation_compare}
\end{table*}

\subsubsection{Simulation Environment}
We utilize the Gazebo simulator to conduct a comparative analysis of OPCD and existing approaches. As the official simulation platform for ROS, Gazebo is renowned for its high-fidelity simulations and has been extensively integrated with various navigation algorithms, thereby facilitating benchmark evaluations. As illustrated in Fig. \ref{warehouse}, a complex 3D environment of 20$m$ $\times$ 20$m$ has been constructed, comprising static buildings (OL1, in purple), operational robots (OL2, in blue), temporarily placed obstacles (OL3, in orange), and pedestrians\footnote{https://github.com/robotics-upo/gazebo\_sfm\_plugin} (OL4, in green). The robot under control, depicted in Fig. \ref{warehouse}(b), is a two-wheeled differential drive robot, whose specifications are consistent with the robot used in the E-Sparrow, as outlined in Section \ref{robot_specification}. Note that OL2 and OL4 are randomly wandering in the environment and will not actively avoid the agent, thus presenting additional challenges for the navigation systems under evaluation and necessitating enhanced safety measures. Furthermore, Rviz\footnote{http://wiki.ros.org/rviz} is employed to visualize the information accessible to the agent, as shown in Fig. \ref{warehouse}(d).

\subsubsection{Evaluation Metrics}
As depicted in Fig. \ref{warehouse}(d), the objective is to navigate the robot from the lower-right corner to the upper-left corner safely and efficiently. To measure the efficiency of the navigation system, two additional metrics were introduced:

\begin{itemize}
\item \textbf{Average distance}: The mean travel distance from the starting point to the endpoint.
\item \textbf{Average time}: The mean travel time from the starting point to the endpoint.
\end{itemize}

It is important to note that these two metrics are only applied to successful cases, as failed cases, which may involve shorter travel distance and time, could distort the overall assessment of navigation efficiency.

\subsubsection{Experiment Results}

The OPCD navigation system is compared with six benchmark navigation systems, as summarized in Table \ref{navigation_compare}. The implementation of these benchmark systems is based on a widely recognized GitHub repository\footnote{https://github.com/ai-winter/ros\_motion\_planning}. To ensure the reliability of the results, each system was evaluated 100 times, and the average results are presented in Table \ref{navigation_compare}. Here, the \textit{Average planning time} metric introduced in Section \ref{Evaluation_ColorDynamic} is excluded from the comparison, as the algorithms are implemented in different programming languages, rendering the comparison of computational time less meaningful.

\begin{figure}
	\centering
	\includegraphics[width=0.48\textwidth]{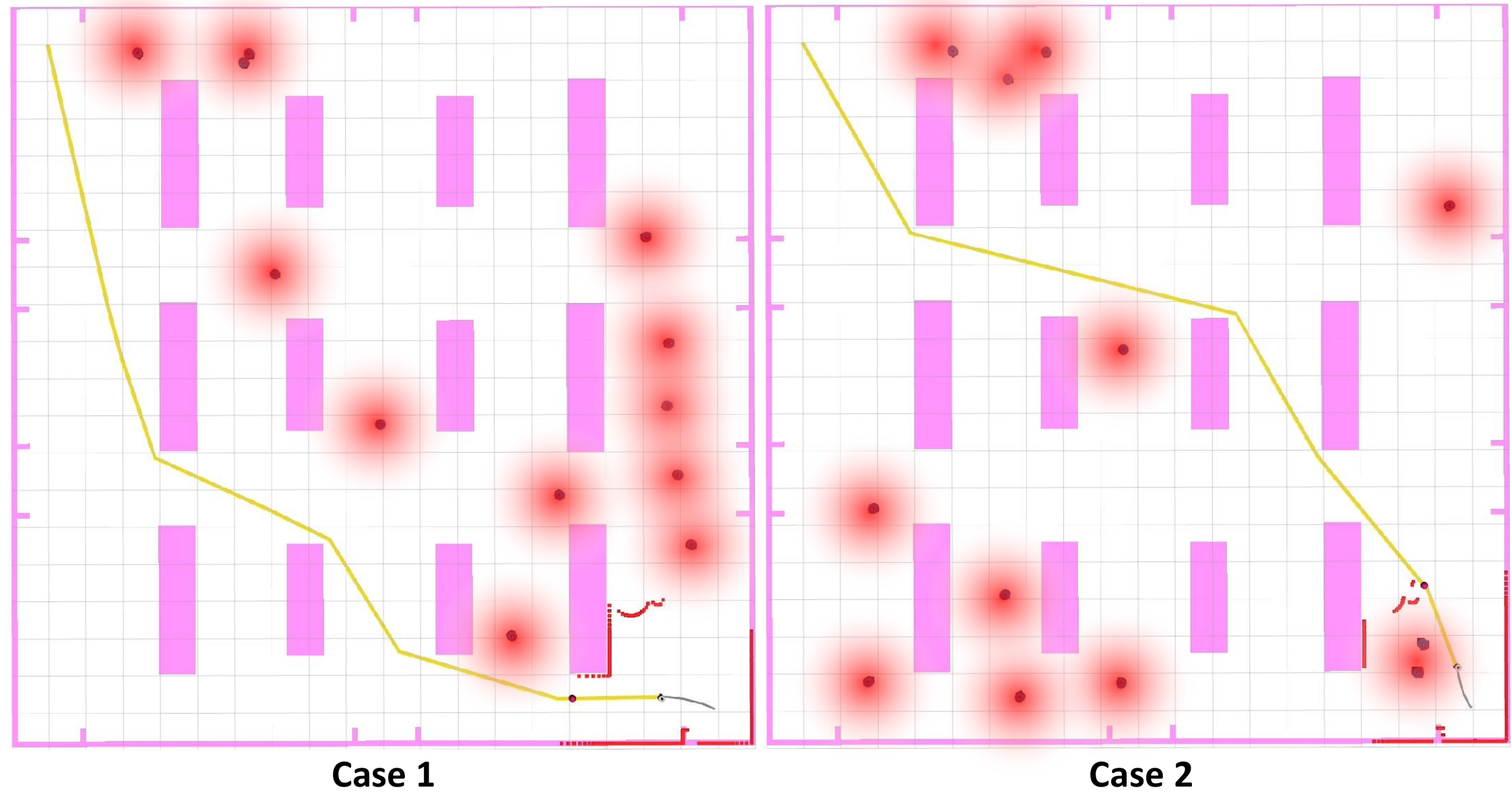}
	\caption{OPCD's proactive circumvention of high-collision-risk (red) areas.}
	\label{dangerous_zoom}
\end{figure}

The numerical results demonstrate that the proposed OPCD system exhibits superior safety in complex scenarios, achieving a 0.96 success rate. This favorable outcome can be attributed to three key features of the OPCD system. First, the global planner, OkayPlan, is capable of avoiding high-collision-risk areas at a global level, thereby enabling the local planner to operate in a safer surrounding, as illustrated in Fig. \ref{dangerous_zoom}. Second, the local planner, ColorDynamic, is able to interpret, predict, and make decisions in complex scenarios, learning high-level strategies that significantly enhance safety, as illustrated in Fig. \ref{skill}. Third, both OkayPlan and ColorDynamic exhibit high computational efficiency. OkayPlan is reported to complete a  planning circle within 8 $ms$\cite{okayplan}, while ColorDynamic in 1.21 $ms$. The real-time performance of OPCD facilitates on-the-fly adjustments in dynamic environments, thereby bolstering navigation safety.


\begin{figure}[t]
	\centering
	\subfloat[]{\includegraphics[width=0.15\textwidth]{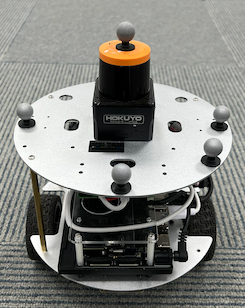}}
	\hfil
	\subfloat[]{\includegraphics[width=0.335\textwidth]{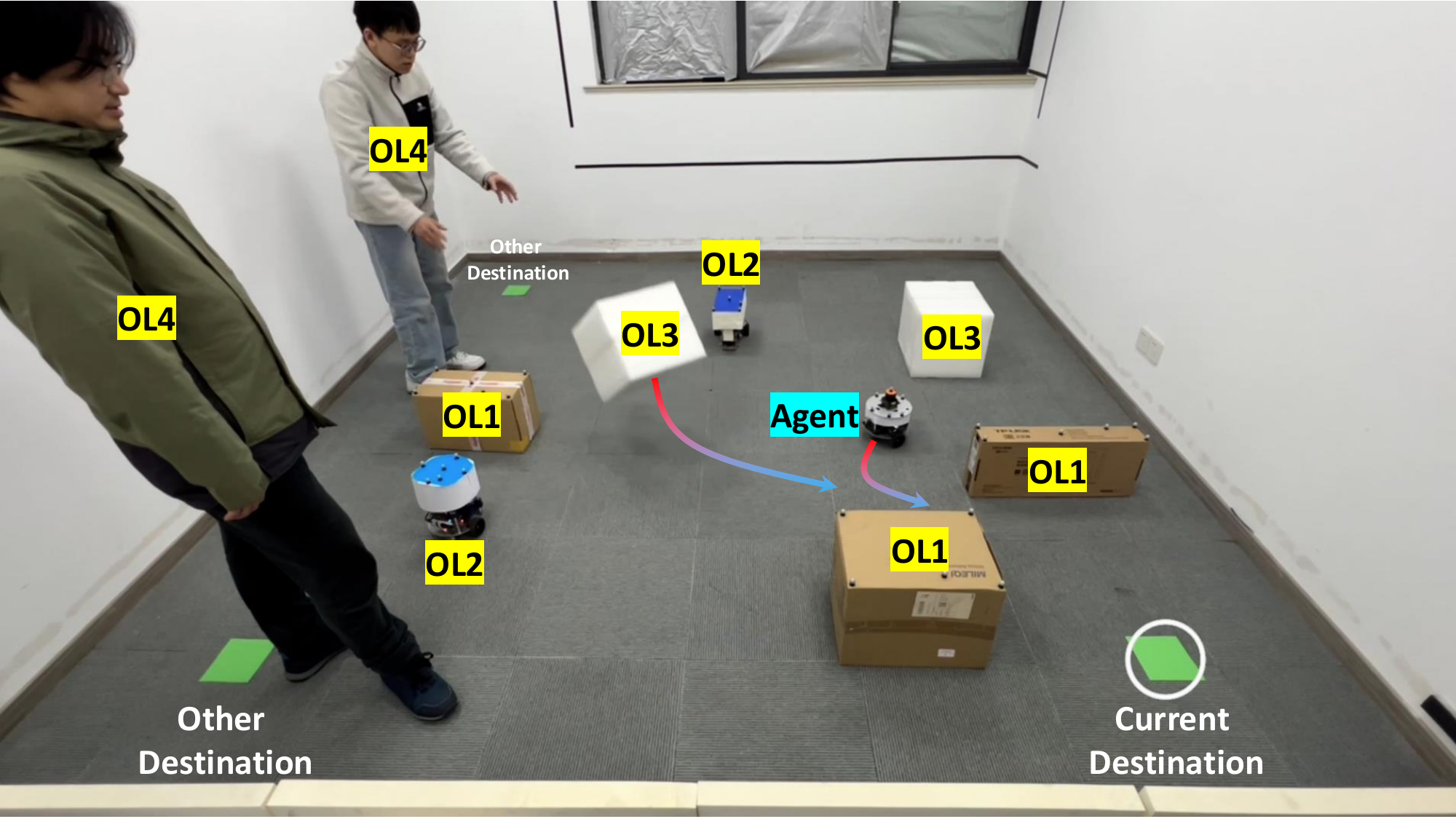}}
	\caption{Real-world (a) platform and (b) environment for OPCD.}
	\label{fml_orin}
\end{figure}

Despite the high success rate, some failure cases remain. During the experiments, it was observed that these failures were predominantly caused by pedestrian oscillation, where pedestrians could abruptly change direction, regardless of physical constraints, and kick the agent. We contend that more realistic pedestrian modeling could mitigate this issue. We have conducted an additional experiment that excludes the pedestrians (only OL1 $\sim$ OL3 remain), wherein the OPCD has achieved a success rate of 0.99.

In terms of navigation efficiency (average distance, time, and speed), OPCD ranks in the middle among the tested systems. This can be attributed to the fact that prioritizing safety may reduce efficiency, as safer navigation typically involves detours. This is evident from the behavior comparison shown in Fig. \ref{behavior}, where OPCD follows a more conservative path, while Systems 4$\sim$6 opt for more aggressive, shorter routes that are riskier.

\subsection{Real-world Deployment of OPCD}

\subsubsection{Real-world Configurations}
This section seeks to evaluate the real-world applicability of the proposed OPCD navigation system. To achieve this, a two-wheeled differential robot, as depicted in Fig. \ref{fml_orin}(a), is employed. The robot is equipped with a \textit{Jetson Orin Nano 8GB} as its onboard computational resource, which runs \textit{ROS Noetic} to facilitate the deployment of the OPCD system. Additionally, the robot is mounted with a \textit{HOKUYO-UST-10LX} lidar, which is responsible for perceiving the surrounding environment and providing the $L(t)$ information necessary for navigation. A motion capture system, FZMotion\footnote{FZMotion Motion Capture System: www.luster3ds.com}, is utilized to extract the kinematic information, $K(t)$, of the robot. Other specifications of the robot are identical to those described in Section \ref{robot_specification}, except that the \textit{HOKUYO} lidar has a 270$^\circ$ field of view. As a result, the local planner is retrained using the ColorDynamic architecture to accommodate this lidar configuration. A real-world environment is constructed as the testing ground for the OPCD system. As illustrated in Fig. \ref{fml_orin}(b), the environment includes predefined destinations, known obstacles (OL1 and OL2), and unknown obstacles (OL3 and OL4). It is important to note that the positions of the known obstacles are also captured via the FZMotion system.

\subsubsection{Experiment Results}
\begin{figure*}
	\centering
	\adjustbox{width=1\textwidth, center}{\includegraphics{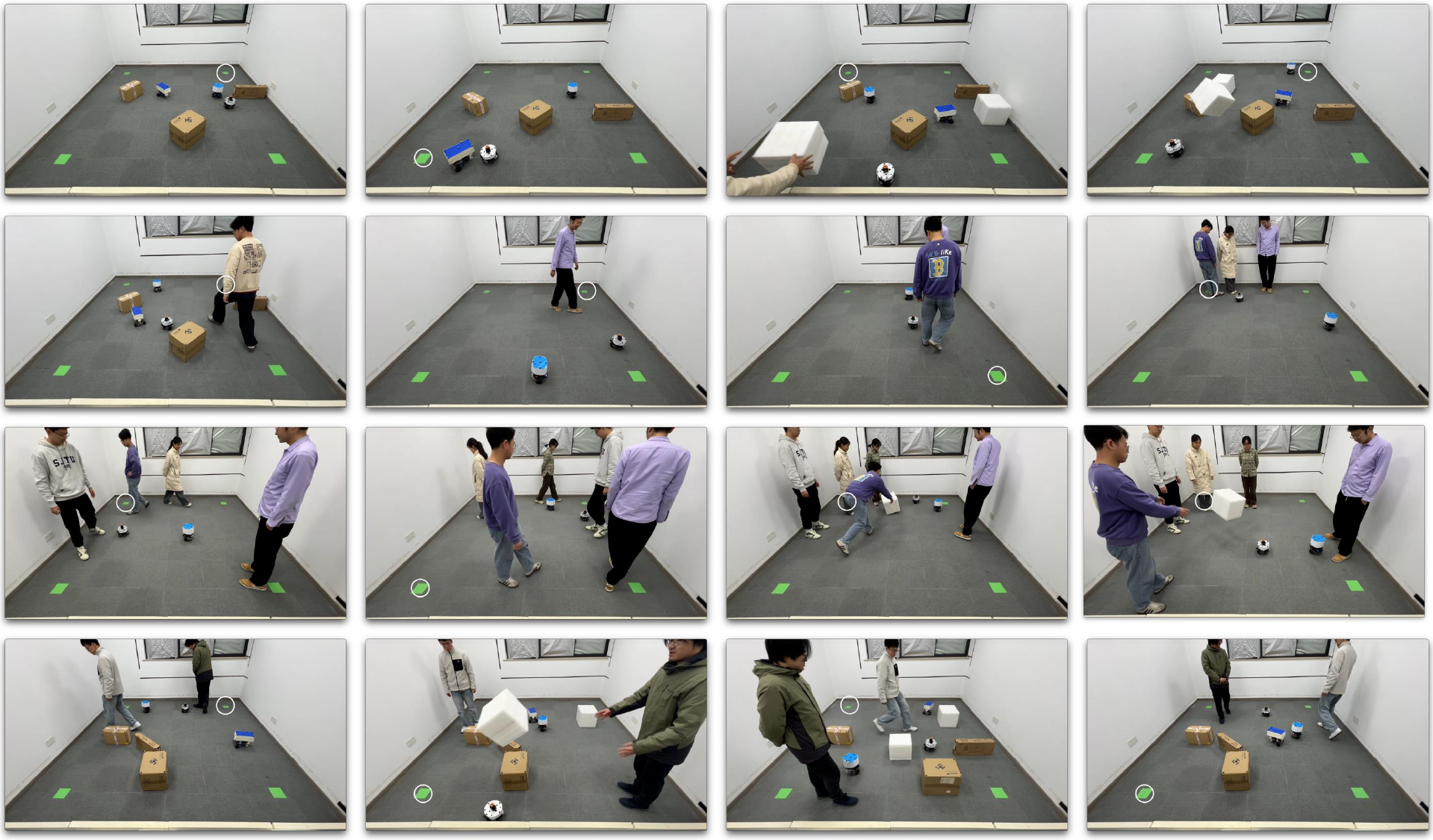}}
	\caption{Snapshots of OPCD’s real-world evaluation. The white robot is controlled by our agent. The brown boxes, blue robots, white boxes, and pedestrians correspond to the OL1, OL2, OL3, and OL4, respectively. The destination for the agent is highlighted by a circle. The video is available on our website\protect\footnotemark[1].}
	\label{realexp}
\end{figure*}

In the evaluation, the OPCD navigation system is tested across a diverse set of map layouts, as shown in the experimental snapshots in Fig. \ref{realexp}. Upon reaching a destination, the agent is randomly assigned another destination and begins a new navigation. The navigation is conducted over 100 trials, with a video of the experiments available on our website\footnotemark[1]. Notably, without any adaptation or fine-tuning in the physical world, the OPCD system achieves a success rate of 0.94. This successful transition from simulation to the real world is ensured by the PGVD. Specifically, by exposing the agent to a wide range of operational conditions during training, a sound generalization ability can be established, allowing the system to handle real-world scenarios as variants of those encountered in training\cite{Color,DR}. The successful deployment of OPCD in highly unstructured environments with dynamic obstacles underscores its effectiveness for mobile robot navigation.

Out of the 100 trials, six failures occur when the agent retreats. These failures are primarily due to the agent’s inability to perceive its rear, as the \textit{HOKUYO} lidar provides only a 270$^\circ$ field of view, leaving a 90$^\circ$ blind spot. It can be anticipated that equipping the robot with a more advanced lidar would mitigate this limitation.

\section{Conclusion}
This study investigates the LPP in highly unstructured environments with dynamic obstacles, introducing a DRL-based solution termed ColorDynamic. 
Specifically, an end-to-end POMDP formulation for LPP is introduced to ensure the compatibility within unstructured environments.
A novel temporal processing network, Transqer, is proposed to enhance comprehension, prediction, and decision-making in dynamic environments.
An efficient simulator, E-Sparrow, featuring PGVD, is developed to facilitate the generation of unlimited yet diverse training data. 
Additionally, a data augmentation technique, referred to as SI, is established to further improve the model’s overall performance. 
Comprehensive experiments, including comparative analyses with state-of-the-art methods, evaluations of generalizability, scalability, and real-time performance, as well as ablation studies, have been conducted to validate the effectiveness of the proposed methods. The results demonstrate that ColorDynamic achieves a success rate exceeding 0.9, with real-time decision-making capabilities (1.2-1.3 $ms$ per planning). 
Furthermore, ColorDynamic is integrated with an efficient global planner, OkayPlan, culminating in the OPCD navigation system. 
Evaluations conducted on the ROS Gazebo platform against off-the-shelf navigation algorithms, alongside real-world deployment, substantiate the superiority and applicability of OPCD. 
Despite promising outcomes, this research is limited by its exclusive focus on 2D navigation. A potential future direction involves extending ColorDynamic and OPCD to 3D navigation, thereby broadening their applicability to unmanned aerial and underwater vehicles.

\section*{Acknowledgments}
We acknowledge the support from the National Natural Science Foundation of China (Grant No. 62273230 and 62203302) and the Oceanic Interdisciplinary Program of Shanghai Jiao Tong University (project number SL2023MS011).
We acknowledge Haodong Yang for his beneficial discussion.
We also acknowledge Qian Liu and Junda Wang for their participation in the real-world experiments.

{\appendix

\begin{table}[htbp]
	\centering
	\caption{Hardware and Software Configuration}
	\begin{tabular}{ll}
		\hline
		\textbf{Hardware} & \textbf{Description}\\
		\hline
		CPU&Intel 10850K\\
		GPU&Nvidia RTX 2070 Super\\
		SSD&Samsumg 990 Pro 1TB \\
		RAM&Kingston 32GB 5600MHz × 2\\
		\hline
		\hline
		\textbf{Software} & \textbf{Description}\\
		\hline
		Computer System&Ubuntu 20.04.1\\
		Robot System&ROS noetic\\
    	Gazebo&11.11.0\\
    	Rviz&1.14.20\\
    	CUDA Driver&550.120\\
    	CUDA Version&12.4\\
		Programming Language&Python 3.8.3\\
		\hline
	\end{tabular}%
	\label{tab:hard_software}%
\end{table}%

}
\begin{figure*}[t]
    \centering
	\adjustbox{width=1.15\textwidth, center}
    {\includegraphics{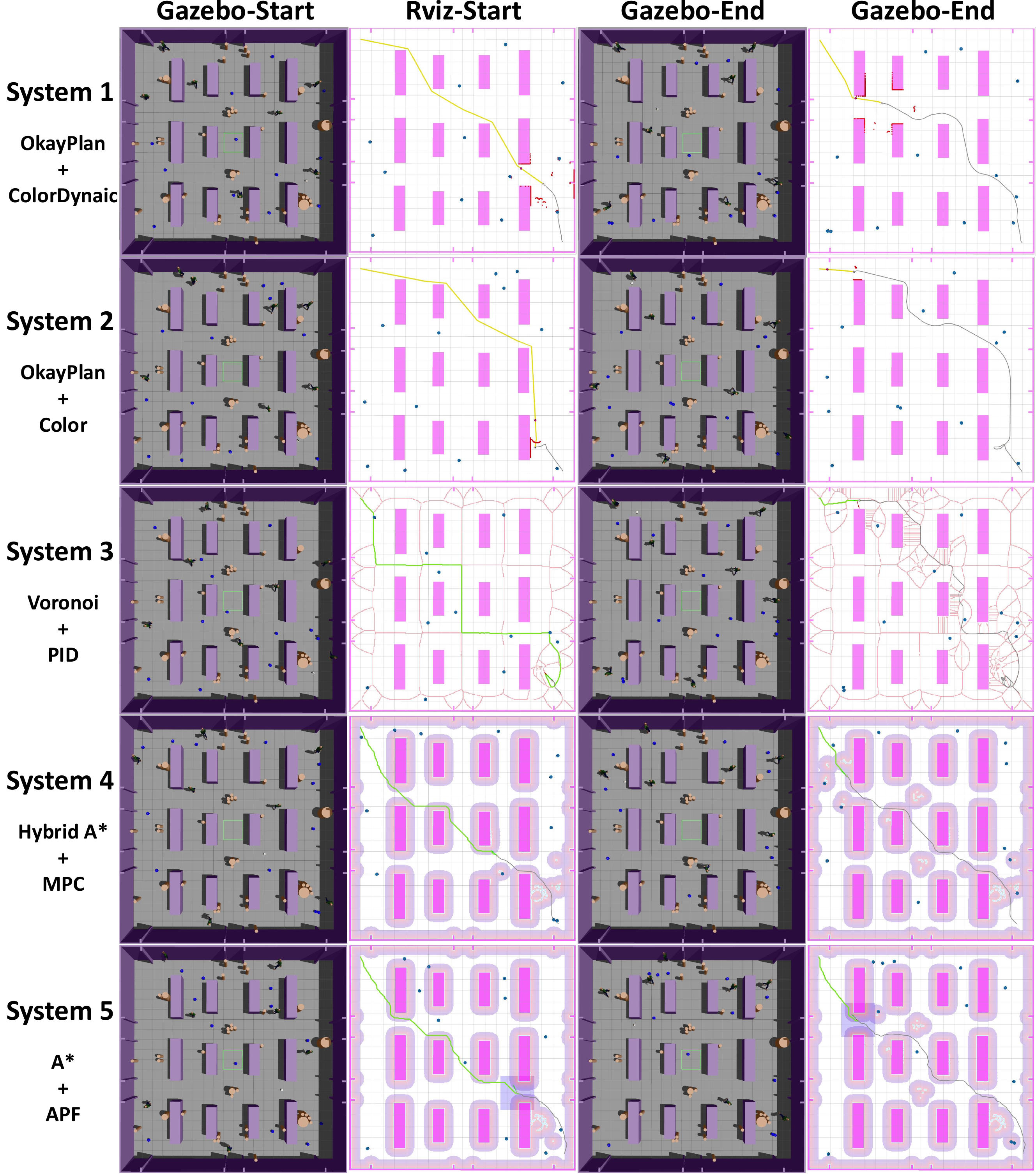}}
    \caption{Behavior comparison of different navigation systems.}
    \label{behavior}
\end{figure*}

\clearpage

\begin{figure*}[t]
	\ContinuedFloat
    \centering
	\adjustbox{width=1.15\textwidth, center}
    {\includegraphics{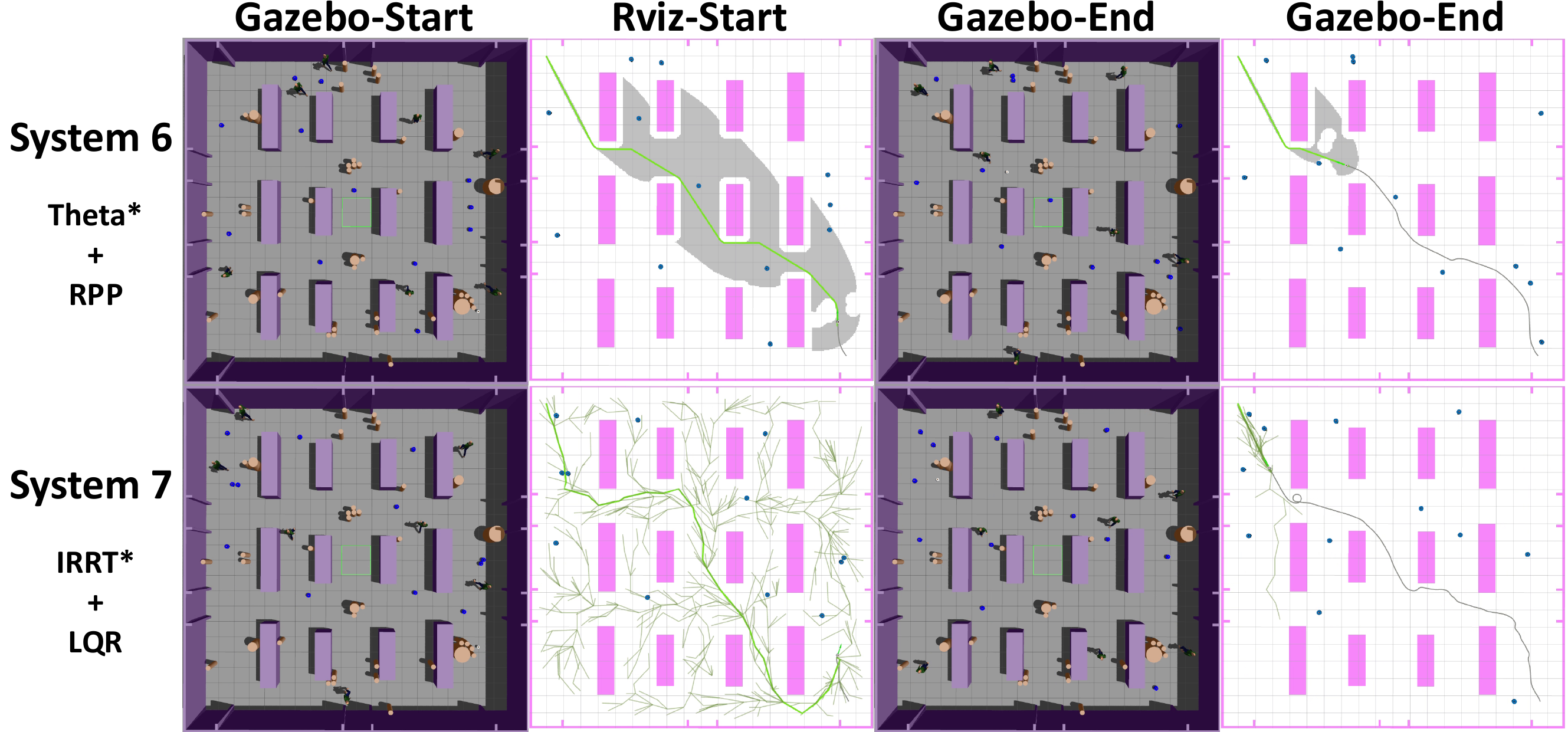}}
    \caption{(Continued) Behavior comparison of different navigation systems.}
\end{figure*}


\bibliographystyle{IEEEtran}
\bibliography{ColorDynamic}

\vfill

\end{document}